\documentclass[runningheads]{llncs}

\usepackage{eccv}

\usepackage{eccvabbrv}

\usepackage{graphicx}
\usepackage{booktabs}
\usepackage{multirow}
\usepackage{wrapfig}
\usepackage[misc]{ifsym}

\usepackage[accsupp]{axessibility}  

\usepackage{hyperref}

\usepackage{orcidlink}

\begin{document}

\title{InteractiveAvatar: Real-Time Streaming Video Generation for Consistent and Intent-Aware Avatars} 

\titlerunning{InteractiveAvatar}

\author{Quanyue Song\inst{1,2}$^{\ast\dagger}$ \and
Yishan He\inst{2}$^{\dagger}$ \and
Yanfei Zhang\inst{2} \and
Shihao Cheng\inst{2,3}$^{\ast}$ \and
Zhixiang He\inst{2} \and
Zhizhi Guo\inst{2}\textsuperscript{\Letter} \and
Chi Zhang\inst{4} \and
Xuelong Li\inst{4} \and
Caigui Jiang\inst{1}\textsuperscript{\Letter}
}

\authorrunning{Q.~Song et al.}

\institute{State Key Laboratory of Human-Machine Hybrid Augmented Intelligence, Institute of Artificial Intelligence and Robotics, Xi'an Jiaotong University, China \and
China Telecom Artificial Intelligence Technology (Beijing) Co., Ltd., China \and
State Key Laboratory of Information Engineering in Surveying, Mapping and Remote Sensing, Wuhan University, China \and
Institute of Artificial Intelligence (TeleAI), China Telecom, China
}

\maketitle
\begingroup
\renewcommand\thefootnote{}\footnotetext{$^{\ast}$ Work done during an internship at China Telecom Artificial Intelligence Technology (Beijing) Co., Ltd.}
\renewcommand\thefootnote{}\footnotetext{$^{\dagger}$ Equal contribution.}
\renewcommand\thefootnote{}\footnotetext{\textsuperscript{\Letter} Corresponding author.}
\endgroup

\begin{abstract}
Recent diffusion-based models have enabled realistic audio-driven avatar generation in real-time streaming. However, existing approaches struggle to maintain visual temporal consistency and fail to explicitly perceive user intent in complex interactive streaming scenarios. To address these challenges, we propose InteractiveAvatar, a real-time infinite-streaming video generation framework that supports visually consistent avatar video generation and intent-aware interactions. With autoregressive distillation, InteractiveAvatar achieves real-time str-eaming generation of human avatars over arbitrarily long durations. For visual consistency, we introduce a Long-Short Visual Memory (LSVM) mechanism that flexibly compresses historical visual information into compact tokens, preserving both short-range coherence and long-term consistency. To generate avatars with speeches and actions aligned with user intent, we propose a Reasoning-Reaction Module (RRM), which incorporates a State-Cycling strategy and a Cache-Switching mechanism. Extensive experimental results over diverse scenarios demonstrate that our method achieves state-of-the-art visual consistency in long-duration generation, while enabling complex user-avatar interaction in real time.
\end{abstract}

\section{Introduction}
With the development of diffusion-based models for audio-driven avatar generation, it has become possible to synthesize videos with accurate lip synchronization, highly realistic and visually appealing appearances~\cite{an2025ai,kong2025let,wang2025fantasytalking,cheng2026unisonharmonizingmotionspeech}. Furthermore, recent advances~\cite{yin2025slow,low2025talkingmachines,sun2025streamavatar} show promising results in real-time streaming diffusion generation of human avatars. However, existing approaches~\cite{zhang2023sadtalker,huang2025live,xie2025x,sun2025streamavatar,wang2026flowact} show limitations in user-avatar interaction and visual consistency in streaming generation. These methods are typically restricted to simple interactive scenarios, such as speaking or listening synchronously with the audio, which is limited in understanding the user intent and maintaining consistency in more complicated scenarios. These limitations highlight two fundamental obstacles that continue to hinder the development of truly realistic and interactive human avatars.

The first challenge lies in maintaining temporal consistency during long-duration video generation. In interactive scenarios with sustained user engagement, visual content evolves continuously, resulting in substantial scene variations and increased complexity. Most existing real-time streaming methods~\cite{huang2025live,sun2025streamavatar} employ causal attention, limiting the model’s receptive field to a few previously generated chunks along with reference image. When the generated content gradually diverges from this initial reference, inconsistencies tend to emerge between adjacent chunks, causing significant temporal drift and the degradation of long-range video consistency.

The second challenge concerns the user-avatar interaction. Beyond generating verbal responses, a realistic avatar is expected to perform actions that are semantically aligned with the user’s intent. For instance, when a user asks "Can you check what time it is on the watch?", the avatar should glance at a watch while verbally providing the time. Existing approaches~\cite{xie2025x,wang2026flowact}, however, predominantly follow an audio-driven paradigm, using response speech generated by large language models to animate lip movements and synthesizing limited actions or gestures based on coarse audio-motion correlations learned from training data. This design overlooks explicit modeling of user intent, making it difficult to generate fine-grained and intent-aware actions such as checking time using a watch, thereby limiting the realism of interactive behavior.

\begin{figure*}[t]
\centering
\includegraphics[width=\textwidth, trim=3 200 3 5, clip]{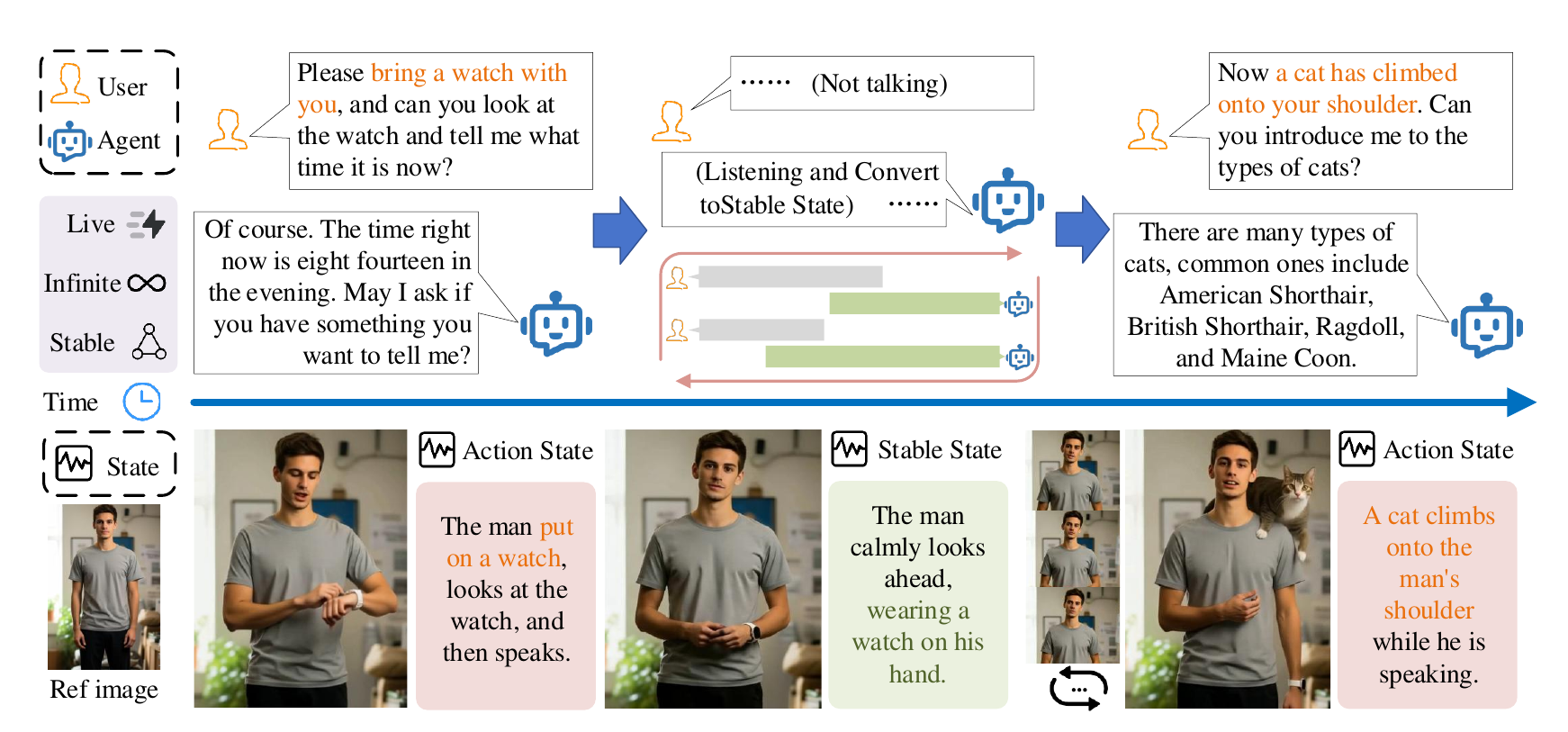} 
\caption{We propose InteractiveAvatar, a real-time streaming audio-driven avatar generation framework that enables intent-aware interaction. InteractiveAvatar interprets user intent to generate contextually relevant actions throughout the dialogue while maintaining long-range visual consistency.}
\label{fig_teaser}
\end{figure*}

To address these challenges, we propose InteractiveAvatar (see Fig.~\ref{fig_teaser}). InteractiveAvatar introduces a real-time interactive generation paradigm, in which the avatar can respond to users by not only producing speech, but also performing intent-aligned actions and adaptively updating the surrounding scene, while maintaining strong visual consistency over long-duration video generation. Specifically, our framework incorporates a Long-Short Visual Memory (LSVM) mechanism and a Reasoning-Reaction Module (RRM) into the real-time streaming diffusion generation.

The LSVM mechanism maintains temporal consistency in streaming generation by jointly modeling short-term and long-term visual information. The short-term memory preserves recently generated frames to ensure coherent transitions and local continuity, while the long-term memory retains representative visual states that capture critical information throughout the entire generation process. We propose a Dynamic Key-Frame Selection strategy, which flexibly transfers visually important content from the short-term memory to the long-term memory during inference, enabling the model to preserve essential global information and prevent temporal drift over long-duration generation.

The RRM enhances the realism of user-avatar interaction by leveraging a large language model for intent understanding and state management. Given user input, it infers the underlying intent and generates corresponding verbal responses and action instructions to guide avatar synthesis. To ensure a natural interaction flow, we design a State-Cycling strategy that enables smooth transitions between listening and execution states. Additionally, memory augmentation is incorporated to maintain consistency between avatar behavior and scene evolution. We further introduce a Cache-Switching mechanism to accelerate action execution and scene transitions, thereby reducing perceived latency during interaction.

By integrating these two components with streaming generation approach, InteractiveAvatar enables real-time synthesis of highly consistent and interactive human avatars. We evaluate InteractiveAvatar across multiple user-instruction scenarios to assess its responsiveness and generation quality. Both qualitative and quantitative results demonstrate that InteractiveAvatar achieves real-time avatar generation while maintaining strong visual consistency over long-duration video sequences, as well as supporting realistic and complex user-avatar interactions.

Contributions of our InteractiveAvatar are listed below:
\begin{itemize}
\item We propose InteractiveAvatar, a novel real-time streaming audio-driven avatar generation framework that supports long-duration video synthesis with strong visual consistency and intent-aware user-avatar interaction.
\item We introduce a Long-Short Visual Memory mechanism that flexibly preserves historical visual representations, enabling the model to maintain attention to past content and significantly improving visual coherence in long-duration generation.
\item We design a Reasoning-Reaction Module equipped with a State-Cycling strategy and a Cache-Switching mechanism, enabling intent-aware interaction and efficient command execution with reduced latency.
\end{itemize}

\section{Related Work}
\subsection{Audio-Driven Avatar Generation}
Audio-driven avatar generation aims to synthesize realistic human videos conditioned on speech while preserving identity and motion consistency. Early methods, such as Wav2Lip~\cite{prajwal2020lip}, focus primarily on accurate lip synchronization but fail to model head and body movements. Later two-staged animators~\cite{zhang2023sadtalker,xu2024vasa} improve controllability on expressions and head movements by first predicting latent motion representations from audio and further rendering avatar expressions and head motions using these latent representations, yet these methods are still constrained by predefined motion priors and lack of natural body movements. More recently, diffusion-based frameworks, particularly DiT-style architectures~\cite{wang2025fantasytalking,kong2025let,chen2025echomimic,cui2025hallo3,tian2024emo,chen2025hunyuanvideo,gan2025omniavatar,meng2025echomimicv3}, have significantly improved audio-driven avatar generation in visual fidelity and naturalness. However, their iterative denoising process incurs substantial computational overhead, hindering real-time streaming and long-term consistency in interactive scenarios. Moreover, most methods rely solely on audio signals, limiting their ability to model high-level user intent.

\subsection{Streaming Video Generation}
Typical bidirectional attention-based video diffusion generators~\cite{chen2025hunyuanvideo,wan2025wan} suffer from low computational efficiency, limited video duration, and poor temporal stability. To address these problems, recent works~\cite{huang2025live,xie2025x,sun2025streamavatar} distill bidirectional diffusion transformers into causal autoregressive architectures for streaming inference. For example, CausVid~\cite{yin2025slow} introduces block-causal attention with distribution matching distillation. And Self-Forcing~\cite{huang2025self} further mitigate train-test mismatches by conditioning on previously generated frames during training and further improve long-horizon stability. Other approaches~\cite{cui2025self,yang2025longlive} explore rolling-window denoising, attention sinks, or KV-recache strategies to reduce temporal drift and error accumulation. Despite these advances, existing methods mainly rely on a limited number of previously generated chunks and focus mainly on local temporal context. This limitation prevents effective modeling of  long-range information across the generation horizon, resulting in degraded long-range consistency and temporal drift.

\subsection{Interactive Avatar Generation}
Interactive avatar generation aims to create avatars capable of engaging in natural and responsive user interactions. Early methods~\cite{zhou2020makelttalk,prajwal2020lip,ginosar2019learning} rely on generating actions conditioned from audio cues, where a speech model drives lip movements and simple gestures. The generated results are largely depend on acoustic cues, with synchronized but simple gestures. Recent works~\cite{low2025talkingmachines,ding2025mtvcrafter} on interactive avatars have explored audio-driven streaming diffusion to enable responsive user-avatar interactions. However, these methods lack multi-turn conversational memory or action-level reasoning. More recent studies~\cite{wang2026flowact,ng2022learning,wang2025diffusion} have incorporated limited “listening states” or predefined actions to simulate interaction, yet they cannot maintain long-term context or produce intent-aware behaviors. In contrast, our approach integrates a human-imitating reasoning module with memory-enhanced diffusion-based video synthesis, enables real-time, context-aware, and temporally consistent interactive avatar behavior, bridging the gap between simple reactive systems and fully intent-aware digital humans.

\begin{figure*}[t]
\centering
\includegraphics[width=\textwidth, trim=20 170 10 5, clip]{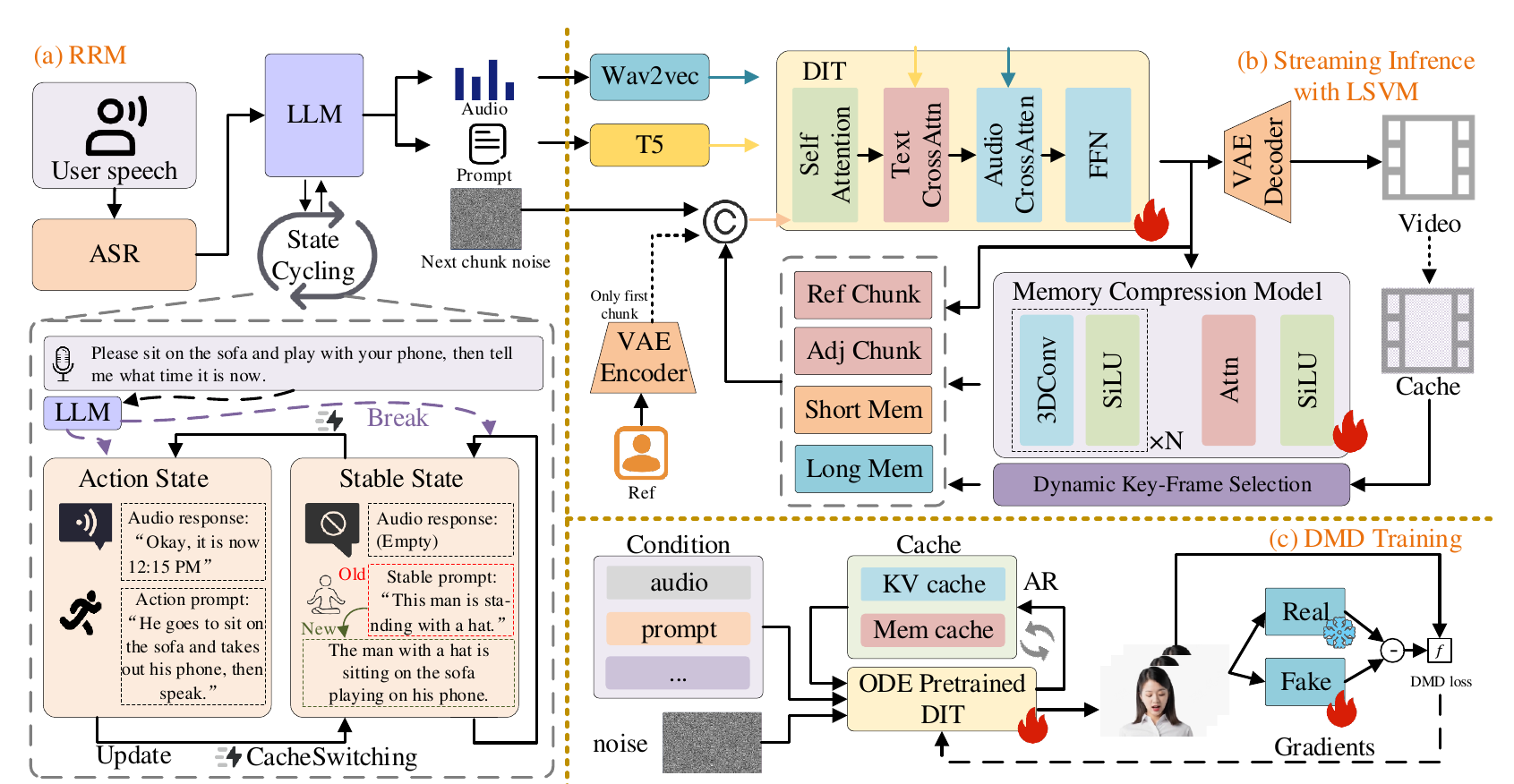} 
\caption{Overview of InteractiveAvatar, which consists of (a) The Reasoning-Reaction Module (RRM) performs intent-aware interaction with user; (b) Streaming Inference with Long-Short Visual Memory (LSVM) mechanism to enhance the visual consistency; and (c) DMD training for real-time streaming generation.}
\label{fig_method}
\end{figure*}

\section{Method}

As illustrated in Figure~\ref{fig_method}, given an avatar reference image and user speech input, InteractiveAvatar aims to generate a real-time video that animates the avatar to naturally respond to the user. The speech signal is first transcribed into text via an automatic speech recognition (ASR) system. The transcribed text is then processed by the Reasoning-Reaction Module (RRM) to produce intent-aware verbal and action instructions (Sec.~\ref{sec:RRM}). Conditioned on these outputs, a streaming video generation model with Long-Short Visual Memory (LSVM) mechanism synthesizes temporally coherent video frames in real time (Sec.~\ref{sec:LSVM}). Finally, the model is trained using self-forcing for autoregressive distillation to enable stable, efficient generation (Sec.~\ref{sec:DMD}), producing intent-aware, temporally consistent avatar videos.

\subsection{Preliminaries}
\subsubsection{Diffusion Transformer}
We build our method upon a Diffusion Transformer (DiT). A pre-trained VAE encodes input data $x$ into latent representations $z = E(x)$. During forward diffusion, Gaussian noise is progressively added to obtain $z_t = \sqrt{\alpha_t}z + \sqrt{1 - \alpha_t}\epsilon$. The DiT model $\epsilon_{\theta}(z_t, t, c)$ predicts the injected noise at timestep $t$ conditioned on signal $c$. Training minimizes the mean squared error between predicted and true noise:
\begin{equation}
    \mathcal{L}=\mathbb{E}_{t, z_{t}, c, \epsilon}\left[\|\epsilon_{\theta}(z_{t}, t, c)-\epsilon\|_2^2\right]
\end{equation}

We adopt Wan2.2 5B~\cite{wan2025wan} as the base model, which employs a causal 3D VAE for spatiotemporal representation learning~\cite{li2022positive,zhang2025variational,li2024survey}. Text conditioning is obtained via T5~\cite{raffel2020exploring} embeddings and injected through cross-attention, while timestep information is incorporated via learned modulation parameters.

\subsubsection{Distribution Matching Distillation}
Distribution Matching Distillation (DMD) distills a pre-trained teacher diffusion model into a few-step student generator by aligning their intermediate noisy distributions~\cite{yin2025slow}. Let $G_\theta(z)$ be the student generator with $\hat{x}=G_\theta(z)$ and $z \sim \mathcal{N}(0,I)$. Denote by $p_{\theta,t}(x_t)$ and $p_{\text{data},t}(x_t)$ the student-induced and teacher-induced distributions at time step $t$, respectively. DMD minimizes the reverse KL divergence:
\begin{equation}
\mathcal{L}_{\text{DMD}}
=
\mathbb{E}_t
\left[
D_{\text{KL}}\big(p_{\theta,t} \,\|\, p_{\text{data},t}\big)
\right].
\end{equation}

Its gradient can be written as:
\begin{equation}
\nabla_\theta \mathcal{L}_{\text{DMD}}=
- \mathbb{E}_{t,z}
\left[\left(
s_{\text{real}}(x_t,t)-s_{\text{fake},\phi}(x_t,t)
\right)^\top
\frac{\partial G_\theta(z)}{\partial \theta}
\right]
\end{equation}
where $x_t=\Psi(\hat{x},t)$, and $s_{\text{real}}$ and $s_{\text{fake},\phi}$ denote the teacher and student score functions. Training alternates between updating $s_{\text{fake},\phi}$ and optimizing $G_\theta$. For multi-step distillation, a student rollout is first constructed, and a random intermediate state is used to stabilize training. In our approach, we adopt DMD to distill the diffusion model into a few-step generator, enabling real-time video generation.

\subsection{Long-Short Visual Memory}
\label{sec:LSVM}

\begin{figure}[t]
    \centering
    \includegraphics[width=\textwidth, trim=0 120 0 0, clip]{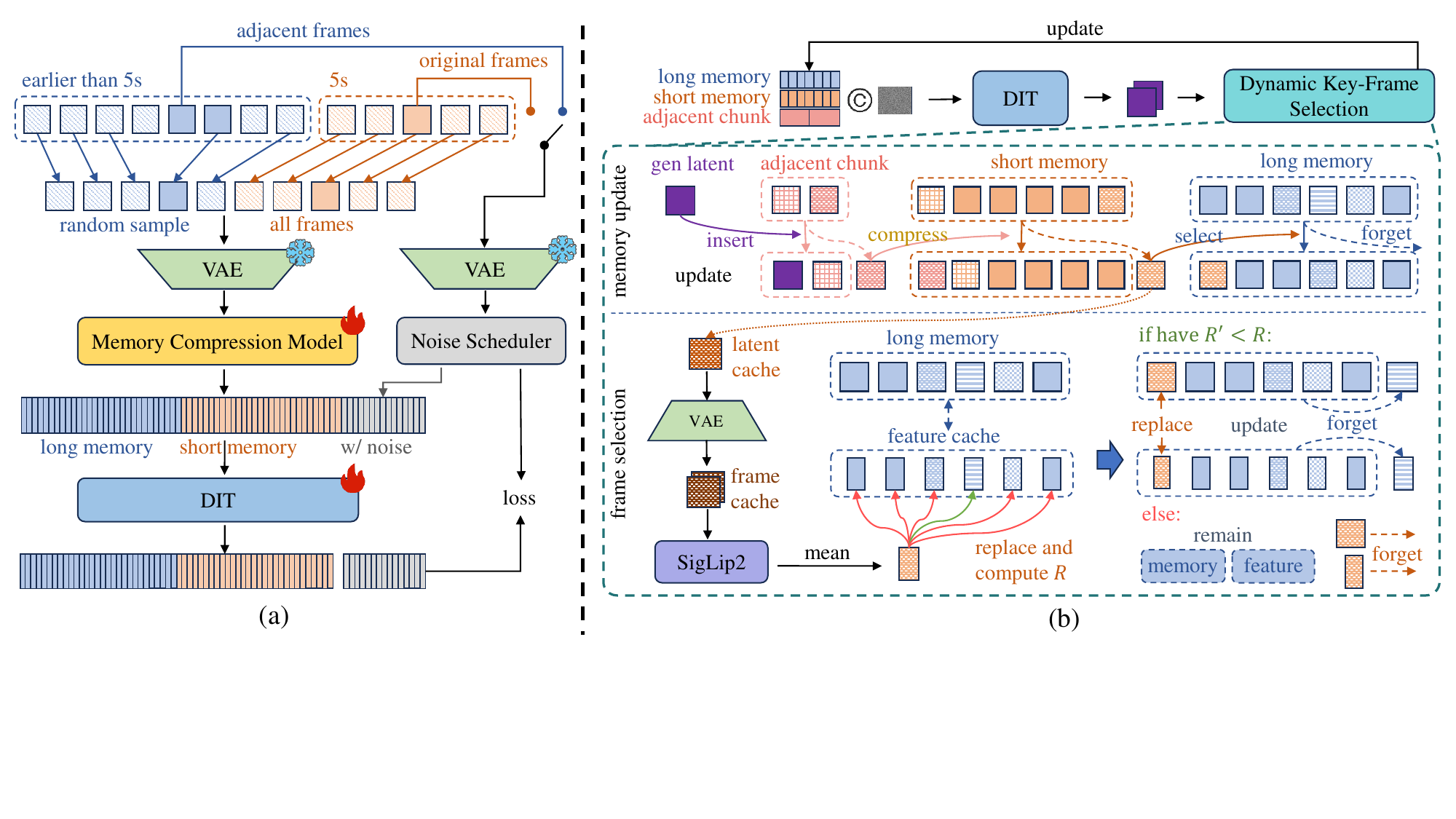}
    \caption{LSVM Mechanism.(a) During training, long-term memory frames are randomly sampled, while short-term memory retains all recent frames. (b) During inference, Dynamic Key-Frame Selection adaptively updates memory to retain critical visual information.}
    \label{fig_memory}
\end{figure}

Preserving key information from previous frames is essential for temporal coherence and visual consistency in long-duration video generation. However, storing all historical frames is computationally expensive and impractical for streaming generation. Inspired by~\cite{zhang2025pretraining} on memory compression and efficient context modeling, we propose Long-Short Visual Memory (LSVM) mechanism (Fig.~\ref{fig_memory}). To effectively capture both global appearance characteristics and recent visual dynamics, we decompose the memory into long-term memory and short-term memory. The short-term memory maintains dense representations of recently generated frames to ensure local temporal coherence, while the long-term memory stores compact representations of globally salient visual states to stabilize overall appearance consistency. A Key-Frame Selection strategy is introduced to determine when short-term memory entries should be promoted to long-term memory, as well as when outdated long-term memory elements should be discarded, enabling adaptive memory updating during streaming generation.

\subsubsection{Long-Short Memory Compression}
We apply a lightweight compression model $\mathcal{C}(\cdot)$ to obtain compact memory tokens. The architecture of $\mathcal{C}(\cdot)$ is similar to~\cite{zhang2025pretraining}, consisting of a sequence of convolutional layers followed by attention modules. Given the generated video latent frames $z_t$, the compact memory tokens can be represented as:
\begin{equation}
m_t = \mathcal{C}(z_t)
\end{equation}
where the temporal compression ratio is set to $1$, ensuring no temporal downsampling during streaming generation. This design avoids temporal merging across latents, which simplifies latent-wise separation and update during streaming generation.

We maintain two memory buffers: a short-term memory $\mathcal{M}_s$ and a long-term memory $\mathcal{M}_l$. The short-term memory stores recent compressed tokens within a fixed window of size $K$ to preserving local continuity:
\begin{equation}
\mathcal{M}_s = \{ m_{t-K+1}, \dots, m_t \}
\end{equation}
The long-term memory maintains a compact set of globally representative tokens with fixed capacity $N$:
\begin{equation}
\mathcal{M}_l = \{ \tilde{m}_1, \dots, \tilde{m}_N \}
\end{equation}

We first concatenate the compressed long-term and short-term memory tokens into a unified history representation $\Phi(H) = \text{Concat}(\mathcal{M}_l, \mathcal{M}_s)$,
where $\mathcal{M}_s$ contains the recent $K$ short-term tokens and $\mathcal{M}_l$ contains the $N$ long-term representative tokens. This design enables the model to jointly capture local temporal continuity and global long-range dependencies.

During training, we randomly sample a video segment $H = \{z_1, \dots, z_T\}$. The last $K$ consecutive frames (we set the length to 5s) of the segment are treated as the short-term memory source $\Omega_s$, while from the preceding history $i < t-K+1$, we randomly sample a fixed number of frame indices $\Omega_l$ to construct the long-term memory:
\begin{equation}
\mathcal{M}_l = \{ m_i \mid i \in \Omega_l,; i < t-K+1 \}
\end{equation}

To train the reconstruction capability, we sample a set of frame indices $\Omega$ from the full segment $H$ as diffusion targets. Depending on the temporal location of each sampled frame, we apply different strategies. For sampled indices within the short-term memory, we randomly keep the subset $\Omega$ unchanged and mask all remaining frames. For sampled indices within the long-term range, we preserve the frame that is temporally closest to $\Omega$ as an anchor frame and mask all remaining frames, as it preserves similar semantic content while providing a slightly different observation for denoising. This design encourages the long-term memory to capture not only pixel-level content but also consistent scene elements.

All masked frames are corrupted using a noise-as-mask strategy. After masking, we clone the clean frames $\{z_i \mid i \in \Omega\}$ as the diffusion targets. The diffusion model is then trained to reconstruct these target frames at arbitrary temporal positions conditioned on the compressed memory representation $\Phi(H)$. This objective can be written as:
\begin{equation}
\mathbb{E}_{H,\Omega,c,\epsilon,t_i}
||
(\epsilon - H_{\Omega})-
G_{\theta}\big((H_{\Omega})_{t_i},t_i,c,\Phi(H)\big)
||_2^2
\end{equation}
where $H_{\Omega}$ denotes the selected clean target frames, $\epsilon_i \sim \mathcal{N}(0, I)$ correspond to sampled noise levels and $\Phi(H)$ is the concatenated long-short memory representation.

By randomly sampling reconstruction targets from both short-term and long-term regions, the model is compelled to encode the entire history in a balanced manner, preserving fine-grained details and global visual consistency essential for stable streaming autoregressive generation.

\subsubsection{Dynamic Key-Frame Selection}
To maintain a compact yet semantically representative long-term memory, we introduce a Dynamic Key-Frame Selection (DKFS) strategy during inference. 

The short-term memory $\mathcal{M}_s$ is implemented as a fixed-length first-in-first-out (FIFO) queue of size $K$. At initialization, all entries are filled with the compressed representation of the first frame $\mathcal{M}_s^{(0)} = \{ m_1, \dots, m_1 \}$. During streaming generation, each newly generated latent frame $z_t$ is compressed into $m_t = \mathcal{C}(z_t)$ and appended to $\mathcal{M}_s$, while the oldest token is removed $\mathcal{M}_s \leftarrow \text{Push}(\mathcal{M}_s, m_t)$. Whenever a token $m_{\text{out}}$ is popped from $\mathcal{M}_s$, the corresponding frame becomes a candidate for long-term memory update.

For each popped frame, we retrieve its corresponding real image frames and feed them into SigLIP2~\cite{tschannen2025siglip} to obtain semantic feature vectors. The features are averaged to produce a single semantic descriptor:
\begin{equation}
\mathbf{s}_{\text{cand}} = \frac{1}{n} \sum_{j=1}^{n} f_{\text{SigLIP2}}(I_j)
\end{equation}
where ${I}_{j}^n$ are the associated real frames and $f_{\text{SigLIP2}}(\cdot)$ denotes the feature extractor.

The long-term memory $\mathcal{M}_l$ is a fixed-capacity buffer of size $N$, initialized by repeating the first frame $\mathcal{M}_l^{(0)} = \{ \tilde{m}_1, \dots, \tilde{m}_1 \}$. Frames selected for long-term memory are inserted in chronological order, ensuring temporal interpretability. To decide whether a candidate frame should be retained in $\mathcal{M}_l$, we evaluate its contribution to global semantic diversity. Let $\{\mathbf{s}_1, \dots, \mathbf{s}_N\}$ denote the current semantic feature set of the long-term memory. For each memory entry $\mathbf{s}_i$, we compute its mean cosine similarity to all other entries and then compute the global redundancy score:
\begin{equation}
\bar{\rho}_i = \frac{1}{N-1} \sum_{j \neq i}
\text{cos}\big(\mathbf{s}_i, \mathbf{s}_j\big),
\quad
R = \frac{1}{N} \sum_{i=1}^{N} \bar{\rho}_i
\end{equation}

If replacing a slot in $\mathcal{M}_l$ with the candidate feature $\mathbf{s}_{\text{cand}}$ leads to a lower redundancy score $R' < R$, the candidate frame is retained and inserted into long-term memory. Otherwise, it is forgot.

This strategy encourages the long-term memory to maintain semantically diverse and globally representative key frames. Instead of storing frames uniformly over time, the memory adaptively preserves frames that introduce novel semantic content, thereby maximizing coverage of distinct scenes, identities, and objects across long video streams.

\subsection{Reasoning-Reaction Module}
\label{sec:RRM}

To enable interactive and controllable avatar behaviors under streaming generation, we introduce a Reasoning-Reaction Module (RRM), which leverages a Large Language Model (LLM)~\cite{xu2025qwen3} for intent understanding and state management. The RRM serves as an information reasoning and response center that bridges multimodal user input and diffusion-based visual generation. It consists of two core components: State-Cycling strategy and a Cache-Switching mechanism.

\subsubsection{State-Cycling Strategy}
Under the State-Cycling strategy, when a user provides speech input, the audio is first transcribed into text and then fed into the LLM together with the current stable state. The LLM outputs two states: an action state, which includes the action prompt $p^{\text{act}}$ and the response audio $a^{\text{resp}}$, and a stable state, which contains the stable-state prompt $p^{\text{stable}}$. The $p^{\text{act}}$ describes the motion to be executed (e.g., standing up or sitting down), the $a^{\text{resp}}$ contains the spoken reply, and the $p^{\text{stable}}$ represents the static visual condition after the action is completed (e.g., the person is sitting calmly). Formally, given user text $x_t$ and previous stable-state prompt $p^{\text{stable}}_{t-1}$, the LLM produces
\begin{equation}
(p^{\text{act}}_t, a^{\text{resp}}_t, p^{\text{stable}}_t)
= \mathrm{LLM}(x_t, \mathcal{S}_{t-1})
\end{equation}

During generation, the diffusion model is conditioned on $(p^{\text{act}}_t, a^{\text{resp}}_t)$ while the response audio is active, where the audio drives lip synchronization and the action prompt guides body motion and pose transitions. Once the response audio finishes, the audio condition is set to empty and the conditioning prompt is replaced by $p^{\text{stable}}_t$. The system then continues generation using only the stable-state prompt, ensuring that the character remains visually consistent without unnecessary motion drift. The model stays in this stable state until a new user instruction arrives and the LLM produces the next action, forming a cyclic process of reasoning, reaction, and stabilization.

\subsubsection{Cache-Switching Mechanism}
To reduce latency under streaming generation, we introduce a Cache-Switching mechanism for prompt-conditioned key-value (KV) cache. Due to the autoregressive setup, previously adjacent generated chunks are computed based on earlier prompts. When the action prompt switches to a new one $p^{\text{act}}_t$, these cached KV become inconsistent with the updated instruction.

To accelerate adaptation to a new prompt, when the action prompt switches, we re-encode the updated text condition and recompute the prompt-conditioned KV tensors of the previously adjacent chunks using the new prompt. These updated KV cache are then used for subsequent attention. Let $\mathcal{K}_{\text{old}}$ denote the cached latent frame KV tensors under the previous prompt $p^{\text{act}}_{t-1}$ and $\mathcal{K}_{\text{new}}$ those recomputed using $p^{\text{act}}_t$. The update is written as
\begin{equation}
\mathcal{K} \leftarrow \mathrm{Replace}(\mathcal{K}_{\text{old}}, \mathcal{K}_{\text{new}}),
\end{equation}
where only the affected chunks are refreshed. This Cache-Switching mechanism enables efficient realignment with updated action prompts, thereby reducing latency during streaming generation.

\subsection{Autoregressive Distillation Adaption}
\label{sec:DMD}

We adapt a pretrained diffusion backbone into a stable and efficient real-time streaming generation model through a four-stage training pipeline. First, we train an image-audio-to-video (AI2V) model based on bidirectional attention on the large-scale audio-visual data, establishing strong speech-driven video generation capability. Second, we pretrain the proposed memory compression module $\mathcal{C}(\cdot)$ on the video reconstruction task so that it can efficiently and adaptively compress visual tokens. Third, we perform ODE initialization by training the model with block-wise causal attention to approximate the bidirectional teacher’s ODE trajectories. The causal attention mask is set as in~\cite{yang2025longlive} during training for both acceleration and high fidelity, along with the optimization of the long short memory compression module. Finally, we apply Self-Forcing DMD training to distill the diffusion model, together with the LSVM, into a few-step autoregressive generator by minimizing the reverse KL divergence between their induced noisy distribution, enabling stable rollout with substantially reduced sampling steps for real-time inference.

\section{Experiments}
\subsection{Experimental Setup}
Our framework is built upon the Wan 2.2 5B~\cite{wan2025wan} model as the backbone. All training and inference are conducted at the resolution of 576p(e.g. 1024x576 in the aspect ratio of 16:9), generating 3 latent per chunk in the streaming setting. Training is performed on 64 NVIDIA H100 GPUs, with 50K steps in Stage 1, 30K steps in Stage 2, and 20K steps in Stage 3 and Stage 4. We employ Fully Sharded Data Parallel (FSDP)~\cite{zhao2023pytorch} with hybrid sharding to reduce memory consumption while enhancing efficiency. The learning rate is set to 1e-5 for the student branch and 2e-6 for the fake score branch. For inference, to enable real-time streaming interation, we first implement kv caching to avoid recomputing key-values of the generated chunks. We also use the pipeline parallelism~\cite{low2025talkingmachines} for further acceleration by deploying the DiT and VAE model on different GPU devices. With these optimizations, we provide an end-to-end latency breakdown to better understand the system bottleneck. Specifically, the DiT module takes approximately 450 ms, the lightweight VAE decoding requires about 50 ms, the LLM audio response accounts for around 1600 ms, and other overhead contributes roughly 450 ms. Taken together, the overall Time-to-First-Frame(TTFF), is approximately 2.6s.

\subsection{Dataset}
We curate a large-scale audio-visual dataset consisting of approximately 3 million high-quality clips after filtering~\cite{jiang2024data}. The dataset is composed of three complementary subsets. The first subset focuses on speech-driven talking-head videos, including HDTF~\cite{zhang2021flow}, VFHQ~\cite{xie2022vfhq}, VoxCeleb2~\cite{Chung18b}, CelebV-Text~\cite{yu2022celebvtext}, and AVSpeech~\cite{zhang2021flow}, which provide high-resolution facial videos with diverse identities and rich speech content for accurate lip synchronization and fine-grained facial motion modeling. The second subset consists of movie and TV-show data primarily collected from OpenHumanVid~\cite{li2024openhumanvid}, capturing complex scenes, expressive performances, and temporal dynamics to improve realism and robustness. The third subset consists of our proprietary conversational dataset, featuring long-duration speaking segments accompanied by rich and diverse body movements. This subset emphasizes sustained speech, expressive gestures, and natural head-shoulder dynamics, providing strong temporal continuity and interaction cues for streaming generation.

\subsection{Metrics}

We evaluate our model from two perspectives: video quality and consistency. For video quality, we assess final video’s perceptual quality using Q-align (IQA)~\cite{wu2023q} and aesthetic appeal (ASE). Distribution-level fidelity is measured by FID~\cite{heusel2017gans} for frame-wise realism and FVD~\cite{unterthiner2018towards} for overall spatio-temporal coherence. For video consistency, we measure audio-visual synchronization using SynC and SynD~\cite{chung2016out}, capturing the correspondence between lip movements and input audio. Object-level temporal consistency (OBJ) is evaluated with Gemini, while identity preservation across frames (ID) is measured using DINOv3~\cite{simeoni2025dinov3}, as avatar videos involve full-body appearance and clothing beyond facial identity. Finally, we use VideoCLIP-XLv2~\cite{wang2024videoclipxladvancinglongdescription} to assesses the alignment between the generated video and the input text prompt (TV), reflecting the effectiveness of semantic control. In addition, we compare the generation speed (FPS) of different models to evaluate real-time performance and streaming efficiency.

\subsection{Results}
We compare InteractiveAvatar against current state-of-the-art open-sourced audio-driven avatar generation approaches, including StableAvatar~\cite{tu2025stableavatarinfinitelengthaudiodrivenavatar}, OmniAvatar~\cite{gan2025omniavatar}, HYAvatar~\cite{chen2025hunyuanvideo}, Hallo3~\cite{cui2025hallo3}, EchoMimicV3~\cite{meng2025echomimicv3}, WanS2V~\cite{wan2025wan} and LiveAvatar~\cite{huang2025live}. We selected 500 videos as our testset and set up interactive scenarios, ranging from simple short to complex long video interactions, with action switches at designated times. For non-real-time models, inference is performed in batches, referencing the last few frames of the previous batch to construct long videos. Quantitative results on relevant objective metrics are presented in Tab.~\ref{tab:comparison}.

\begin{figure*}[tb]
\centering
\includegraphics[width=\textwidth, trim=5 140 5 5, clip]{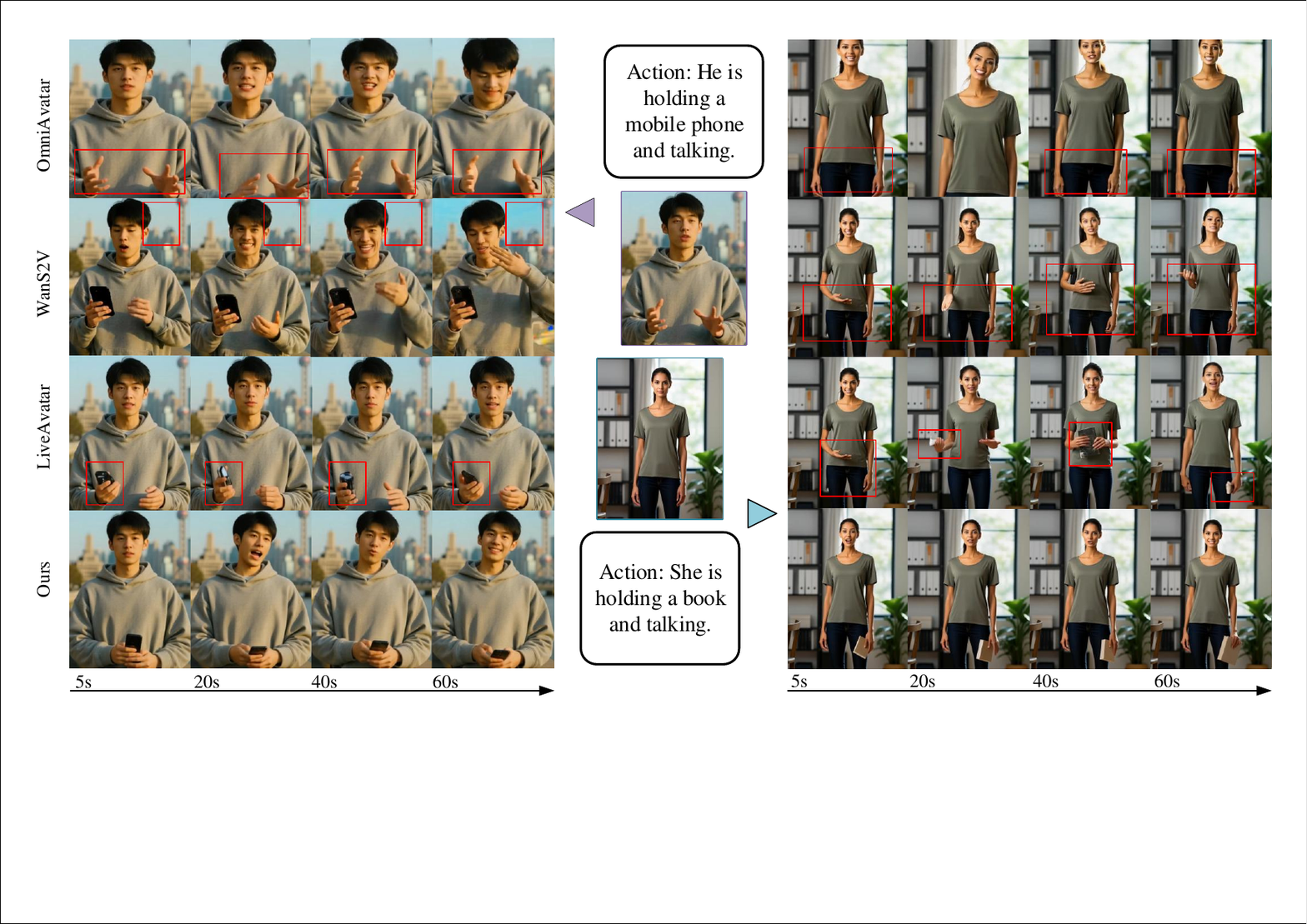} 
\caption{Qualitative comparisons with state-of-the-art methods. Our method exhibits better visual consistency and following of action instructions.}
\label{fig_result}
\end{figure*}

\begin{table}[tb]
\caption{Quantitative comparison with state-of-the-art methods. Best in \textbf{bold} and second best \underline{underlined}. Experiments are conducted using the H100 GPU.}
\label{tab:comparison}
\begin{tabular}{lcccccccccc}
\toprule
\multirow{2}{*}{Model} &
\multicolumn{4}{c}{Video Quality} & \multicolumn{5}{c}{Consistency} & \multicolumn{1}{c}{Speed}\\
\cmidrule(lr){2-5}
\cmidrule(lr){6-10}
\cmidrule(lr){11-11}
&{IQA $\uparrow$} & {ASE $\uparrow$} & {FID $\downarrow$} & {FVD $\downarrow$} & {SynC $\uparrow$} & {SynD $\downarrow$} & {OBJ $\uparrow$} & {ID $\uparrow$} & {TV $\uparrow$} & {FPS $\uparrow$} \\
\midrule
{StableAvatar} & {3.91} & {3.82} & \underline{79.7} & \underline{654.1} & {3.57} & {10.27} & {79.5} & {4.41} & {25.34} & {0.69} \\
{OmniAvatar} & {3.77}  & {3.87} & {94.1} & {831.9} & \textbf{4.94} & \textbf{7.87} & \underline{82.8} & {4.38} & {24.57} & {0.17} \\
{HYAvatar} & {3.81} & \textbf{3.93} & \textbf{76.5} & \textbf{632.6} & {4.78} & {8.11} & {78.9} & {4.46} & {25.61} & {0.09} \\
{Hallo3} & {3.57} & {3.29} & {112.5} & {1127.6} & {4.21} & {9.74} & {75.1} & {4.31} & {24.92} & {0.28} \\
{EchoMimicV3} & \textbf{3.96} & {3.89} & {85.2} & {773.9} & {3.89} & {10.09} & {80.3} & {4.45} & {25.56} & {0.81} \\
{WanS2V} & {3.76} & {3.68} & {88.4} & {793.5} & {4.54} & {8.95} & {82.6} & {4.49} & {25.14} & {0.26} \\
{LiveAvatar} & \underline{3.94} & \underline{3.91} & {83.9} & {672.7} & \underline{4.91} & {8.17} & {76.9} & \textbf{4.53} & \underline{25.78} & \underline{21.94} \\
{Ours} & {3.87} & {3.89} & {80.2} & {701.4} & {4.86} & \underline{7.91} & \textbf{85.2} & \underline{4.51} & \textbf{25.93} & \textbf{26.68}\\
\bottomrule
\end{tabular}
\centering
\smallskip
\end{table}

In terms of video quality, while it does not achieve the best IQA, ASE, FID and FVD scores, InteractiveAvatar maintains reasonable visual fidelity, indicating its ability to generate avatars with consistent and plausible appearance. Regarding temporal and semantic consistency, InteractiveAvatar demonstrates clear advantages. It achieves the highest OBJ and TV, and competitive ID score, indicating strong object consistency across frames, robust identity preservation, and stable temporal variation throughout the video. Moreover, our model maintains accurate lip synchronization and exhibits improved alignment to user instructions compared with other baselines. These results highlight that InteractiveAvatar can generate videos that are not only visually coherent over time but also faithful to both content and user intent. Moreover, by leveraging the lightweight 5B model, InteractiveAvatar reduces hardware requirements while achieving higher inference speed and faster than all other baselines.

Qualitative visualizations further compare our method with OmniAvatar~\cite{gan2025omniavatar}, WanS2V~\cite{wan2025wan} and LiveAvatar~\cite{huang2025live} in Fig.~\ref{fig_result}. In the two demonstrated scenarios, OmniAvatar fails to follow the action instructions and exhibits abrupt camera angle changes. WanS2V shows noticeable quality degradation over time in the first scenario and fails to follow action instructions in the second. LiveAvatar successfully triggers the specified objects according to the action instructions, but the objects’ shapes and colors change rapidly during inference, resulting in low object consistency. In contrast, our method follows the input action instructions while maintaining high consistency for both objects and avatars throughout the interaction.

\subsection{Ablation Studies}

\begin{figure*}[t]
\centering
\includegraphics[width=\textwidth, trim=5 200 5 5, clip]{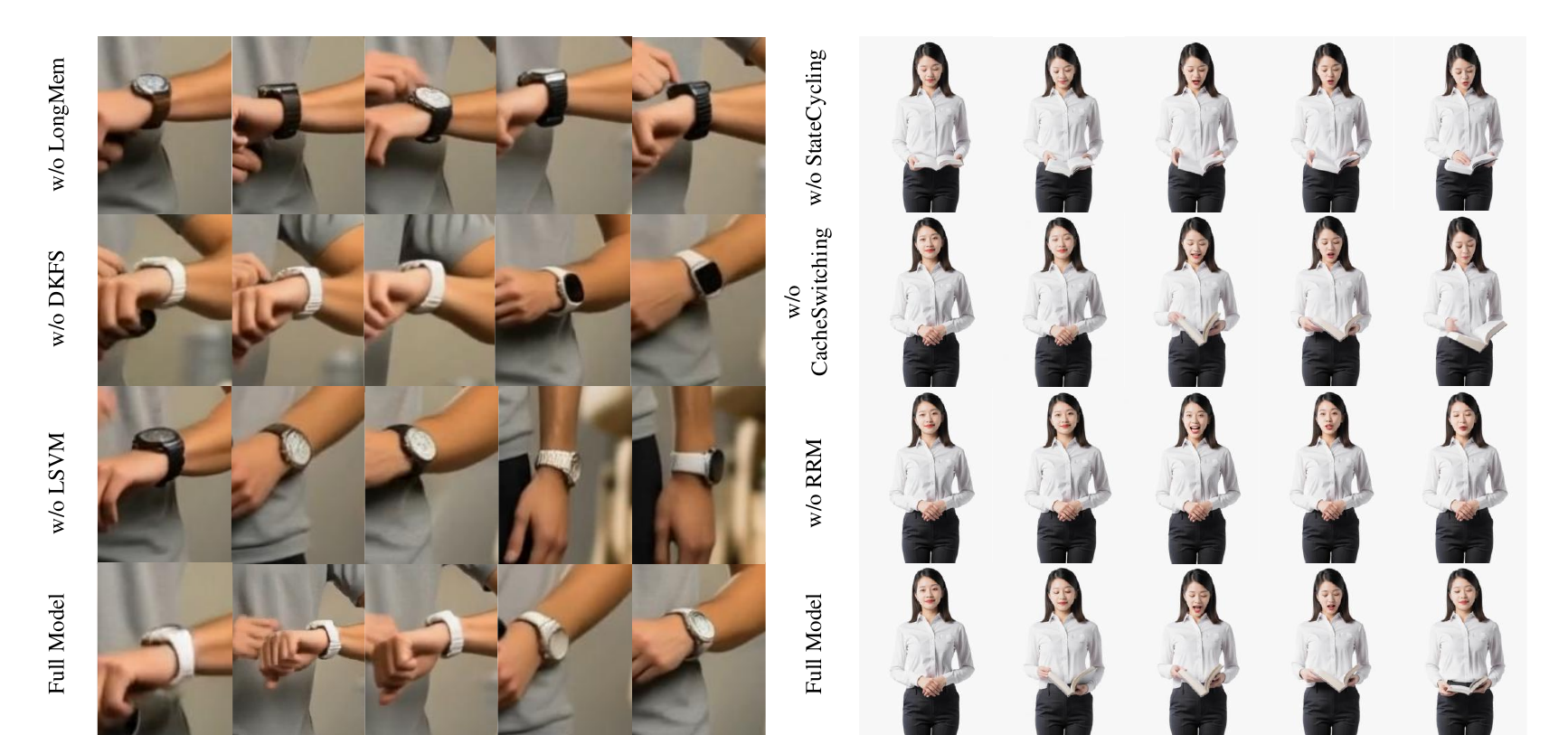} 
\caption{Qualitative ablation of InteractiveAvatar. Ablation studies show that our Full model maintains the best visual consistency and enables more realistic interactions.}
\label{fig_ablation}
\end{figure*}

We conduct ablation studies to analyze the impact of each component in our framework by systematically modifying or removing key modules, as summarized in Table~\ref{tab:ablation}. We evaluate both consistency-related metrics (OBJ, ID, TV) and inference speed (FPS).

\setlength{\intextsep}{0pt} 
\begin{wraptable}{r}{0.60\textwidth} 
\centering
\caption{Ablation on the LSVM, RRM and DMD.\\[10pt]}
\label{tab:ablation}
\begin{tabular}{lcccc}
\toprule
Method & {OBJ $\uparrow$} & {ID $\uparrow$} & {TV $\uparrow$} & {FPS $\uparrow$}\\
\midrule
w/o LongMem & 82.6 & 4.43 & \underline{25.91} & \underline{28.92}\\
w/o DKFS & 83.1 & 4.45 & 25.85 & 26.83\\
w/o LSVM & 78.4 & 4.38 & 25.87 & \textbf{30.04}\\
\midrule
w/o StateCycling & 84.1 & 4.46 & 25.42 & 26.68\\
w/o CacheSwitching & \underline{84.5} & \underline{4.47} & 25.76 & 26.75\\
w/o RRM & 83.8 & 4.46 & 24.89 & 26.75\\
\midrule
w/o DMD & - & - & - & 1.27\\
\midrule
Ours & \textbf{85.2} & \textbf{4.51} & \textbf{25.93} & {26.68}\\
\bottomrule
\end{tabular}
\\[4pt]
\end{wraptable}

For the ablation study of the LSVM module, we design a scenario where the avatar wears a watch during interaction. Removing the long-term memory token (w/o LongMem) degrades object and identity consistency, demonstrating the necessity of long-range temporal modeling. Replacing Dynamic Key-Frame Selection with random sampling (w/o DKFS) causes slight distortions in the watch face, highlighting the advantage of informed memory updates. Removing the entire LSVM module (w/o LSVM) leads to a significant drop in OBJ, confirming its importance for visual consistency, though FPS increases due to reduced computation. As shown in Fig.~\ref{fig_ablation}, only the full model reliably preserves the watch’s shape and color.

For the Reasoning-Reaction Module (RRM), removing the State Cycling strategy and using a fixed action prompt (w/o StateCycling) causes the model to repeatedly execute a single action, reducing interaction realism. Disabling Cache-Switching (w/o CacheSwitching) increases response latency to prompt changes, which can cause delays or failures in the onset of prompt-related motions, making actions such as picking up and opening a book noticeably slower. Removing the entire RRM (w/o RRM) and reverting to a default prompt prevents the avatar from understanding user intent, resulting in only verbal responses without corresponding actions. These results demonstrate the necessity of dynamic state control and intent-aware reasoning. In contrast, our full model achieves natural interaction with low latency.

Finally, inference speed drops dramatically without DMD distillation (w/o DMD), verifying that DMD is essential for real-time performance. We do not report other metrics for the w/o DMD setting, as the model fails to achieve long video generation. When inference is performed by using the last frame of the previous segment as the first frame of the next segment, the video quality progressively degrades after multiple segments.

\section{Conclusion}
In this work, we present InteractiveAvatar, a novel real-time streaming diffusion framework that bridges the gap between high-quality avatar generation and realistic, intent-aware human-avatar interaction by jointly addressing two core challenges, long visual consistency and interactive responsiveness. Specifically, the proposed LSVM mechanism effectively mitigates temporal drift during extended streaming generation, preserving both local coherence and global identity consistency. Meanwhile, the RRM empowers the avatar to interpret user intent and generate coordinated speech and actions. In summary, InteractiveAvatar achieves real-time, long-duration, and intent-aware avatar generation, providing a practical foundation for next-generation immersive and intelligent digital humans.

\subsubsection{Limitations and Future work} Due to the limited memory capacity and key-frame selection strategy, long videos or large motions, especially those that deviate significantly from the reference frame, may degrade as early latents are discarded, or some generated objects may gradually disappear. In addition, objects absent from the first frame may appear abruptly when mentioned in the instruction, which is partly determined by the base model capability. Although prompt design can help smooth object emergence, this remains challenging. In future work, we aim to train end-to-end generative models with larger-scale data and computation to achieve smoother and more natural avatar interaction.



\section*{Acknowledgements}
We thank our colleagues for their assistance with data processing, and the anonymous reviewers for their suggestive comments. This paper is supported by NSFC under grant No. 62495092, No.62125305 and Natural Science Basic Research Plan in Shaanxi Province of China (No. 2025SYS-SYSZD-023).

%
%
\bibliographystyle{splncs04}
\bibliography{main}

@String(CVPR  = {IEEE Conf. Comput. Vis. Pattern Recog.})

@String(CVPRW = {IEEE Conf. Comput. Vis. Pattern Recog. Worksh.})

@String(AAAI  = {AAAI})

@String(TOG   = {ACM Trans. Graph.})

@String(CVPR  = {CVPR})

@String(CVPRW = {CVPRW})

@String(TOG   = {ACM TOG})

@article{an2025ai,
  title={Ai flow: Perspectives, scenarios, and approaches (2025)},
  author={An, Hongjun and Hu, Wenhan and Huang, Sida and Huang, Siqi and Li, Ruanjun and Liang, Yuanzhi and Shao, Jiawei and Song, Yiliang and Wang, Zihan and Yuan, Cheng and others},
  journal={arXiv preprint arXiv:2506.12479},
  year={2025}
}

@article{li2022positive,
  title={Positive-incentive noise},
  author={Li, Xuelong},
  journal={IEEE Transactions on Neural Networks and Learning Systems},
  volume={35},
  number={6},
  pages={8708--8714},
  year={2022},
  publisher={IEEE}
}

@article{zhang2025variational,
  title={Variational positive-incentive noise: How noise benefits models},
  author={Zhang, Hongyuan and Huang, Sida and Guo, Yubin and Li, Xuelong},
  journal={IEEE Transactions on Pattern Analysis and Machine Intelligence},
  year={2025},
  publisher={IEEE}
}

@article{li2024survey,
  title={A survey on LLM-based multi-agent systems: workflow, infrastructure, and challenges},
  author={Li, Xinyi and Wang, Sai and Zeng, Siqi and Wu, Yu and Yang, Yi},
  journal={Vicinagearth},
  volume={1},
  number={1},
  pages={9},
  year={2024},
  publisher={Springer}
}

@article{jiang2024data,
  title={Data augmentation in human-centric vision},
  author={Jiang, Wentao and Zhang, Yige and Zheng, Shaozhong and Liu, Si and Yan, Shuicheng},
  journal={Vicinagearth},
  volume={1},
  number={1},
  pages={8},
  year={2024},
  publisher={Springer}
}

@inproceedings{wang2025fantasytalking,
  title={Fantasytalking: Realistic talking portrait generation via coherent motion synthesis},
  author={Wang, Mengchao and Wang, Qiang and Jiang, Fan and Fan, Yaqi and Zhang, Yunpeng and Qi, Yonggang and Zhao, Kun and Xu, Mu},
  booktitle={Proceedings of the 33rd ACM International Conference on Multimedia},
  pages={9891--9900},
  year={2025}
}

@article{kong2025let,
  title={Let them talk: Audio-driven multi-person conversational video generation},
  author={Kong, Zhe and Gao, Feng and Zhang, Yong and Kang, Zhuoliang and Wei, Xiaoming and Cai, Xunliang and Chen, Guanying and Luo, Wenhan},
  journal={arXiv preprint arXiv:2505.22647},
  year={2025}
}

@inproceedings{yin2025slow,
  title={From slow bidirectional to fast autoregressive video diffusion models},
  author={Yin, Tianwei and Zhang, Qiang and Zhang, Richard and Freeman, William T and Durand, Fredo and Shechtman, Eli and Huang, Xun},
  booktitle={Proceedings of the IEEE/CVF Conference on Computer Vision and Pattern Recognition},
  pages={22963--22974},
  year={2025}
}

@article{huang2025live,
  title={Live avatar: Streaming real-time audio-driven avatar generation with infinite length},
  author={Huang, Yubo and Guo, Hailong and Wu, Fangtai and Zhang, Shifeng and Huang, Shijie and Gan, Qijun and Liu, Lin and Zhao, Sirui and Chen, Enhong and Liu, Jiaming and others},
  journal={arXiv preprint arXiv:2512.04677},
  year={2025}
}

@article{xie2025x,
  title={X-streamer: Unified human world modeling with audiovisual interaction},
  author={Xie, You and Gu, Tianpei and Li, Zenan and Zhang, Chenxu and Song, Guoxian and Zhao, Xiaochen and Liang, Chao and Jiang, Jianwen and Xu, Hongyi and Luo, Linjie},
  journal={arXiv preprint arXiv:2509.21574},
  year={2025}
}

@article{sun2025streamavatar,
  title={StreamAvatar: Streaming Diffusion Models for Real-Time Interactive Human Avatars},
  author={Sun, Zhiyao and Peng, Ziqiao and Ma, Yifeng and Chen, Yi and Zhou, Zhengguang and Zhou, Zixiang and Zhang, Guozhen and Zhang, Youliang and Zhou, Yuan and Lu, Qinglin and others},
  journal={arXiv preprint arXiv:2512.22065},
  year={2025}
}

@article{wang2026flowact,
  title={FlowAct-R1: Towards Interactive Humanoid Video Generation},
  author={Wang, Lizhen and Zhu, Yongming and Ge, Zhipeng and Zheng, Youwei and Zhang, Longhao and Hu, Tianshu and Qin, Shiyang and Luo, Mingshuang and Zhang, Jiaxu and Chen, Xin and others},
  journal={arXiv preprint arXiv:2601.10103},
  year={2026}
}

@article{low2025talkingmachines,
  title={Talkingmachines: Real-time audio-driven facetime-style video via autoregressive diffusion models},
  author={Low, Chetwin and Wang, Weimin},
  journal={arXiv preprint arXiv:2506.03099},
  year={2025}
}

@inproceedings{zhang2023sadtalker,
  title={Sadtalker: Learning realistic 3d motion coefficients for stylized audio-driven single image talking face animation},
  author={Zhang, Wenxuan and Cun, Xiaodong and Wang, Xuan and Zhang, Yong and Shen, Xi and Guo, Yu and Shan, Ying and Wang, Fei},
  booktitle={Proceedings of the IEEE/CVF conference on computer vision and pattern recognition},
  pages={8652--8661},
  year={2023}
}

@inproceedings{prajwal2020lip,
  title={A lip sync expert is all you need for speech to lip generation in the wild},
  author={Prajwal, KR and Mukhopadhyay, Rudrabha and Namboodiri, Vinay P and Jawahar, CV},
  booktitle={Proceedings of the 28th ACM international conference on multimedia},
  pages={484--492},
  year={2020}
}

@article{xu2024vasa,
  title={Vasa-1: Lifelike audio-driven talking faces generated in real time},
  author={Xu, Sicheng and Chen, Guojun and Guo, Yu-Xiao and Yang, Jiaolong and Li, Chong and Zang, Zhenyu and Zhang, Yizhong and Tong, Xin and Guo, Baining},
  journal={Advances in Neural Information Processing Systems},
  volume={37},
  pages={660--684},
  year={2024}
}

@inproceedings{chen2025echomimic,
  title={Echomimic: Lifelike audio-driven portrait animations through editable landmark conditions},
  author={Chen, Zhiyuan and Cao, Jiajiong and Chen, Zhiquan and Li, Yuming and Ma, Chenguang},
  booktitle={Proceedings of the AAAI Conference on Artificial Intelligence},
  volume={39},
  number={3},
  pages={2403--2410},
  year={2025}
}

@inproceedings{cui2025hallo3,
  title={Hallo3: Highly dynamic and realistic portrait image animation with video diffusion transformer},
  author={Cui, Jiahao and Li, Hui and Zhan, Yun and Shang, Hanlin and Cheng, Kaihui and Ma, Yuqi and Mu, Shan and Zhou, Hang and Wang, Jingdong and Zhu, Siyu},
  booktitle={Proceedings of the Computer Vision and Pattern Recognition Conference},
  pages={21086--21095},
  year={2025}
}

@inproceedings{tian2024emo,
  title={Emo: Emote portrait alive generating expressive portrait videos with audio2video diffusion model under weak conditions},
  author={Tian, Linrui and Wang, Qi and Zhang, Bang and Bo, Liefeng},
  booktitle={European Conference on Computer Vision},
  pages={244--260},
  year={2024},
  organization={Springer}
}

@article{chen2025hunyuanvideo,
  title={Hunyuanvideo-avatar: High-fidelity audio-driven human animation for multiple characters},
  author={Chen, Yi and Liang, Sen and Zhou, Zixiang and Huang, Ziyao and Ma, Yifeng and Tang, Junshu and Lin, Qin and Zhou, Yuan and Lu, Qinglin},
  journal={arXiv preprint arXiv:2505.20156},
  year={2025}
}

@article{gan2025omniavatar,
  title={Omniavatar: Efficient audio-driven avatar video generation with adaptive body animation},
  author={Gan, Qijun and Yang, Ruizi and Zhu, Jianke and Xue, Shaofei and Hoi, Steven},
  journal={arXiv preprint arXiv:2506.18866},
  year={2025}
}

@article{meng2025echomimicv3,
  title={Echomimicv3: 1.3 b parameters are all you need for unified multi-modal and multi-task human animation},
  author={Meng, Rang and Wang, Yan and Wu, Weipeng and Zheng, Ruobing and Li, Yuming and Ma, Chenguang},
  journal={arXiv preprint arXiv:2507.03905},
  year={2025}
}

@article{huang2025self,
  title={Self forcing: Bridging the train-test gap in autoregressive video diffusion},
  author={Huang, Xun and Li, Zhengqi and He, Guande and Zhou, Mingyuan and Shechtman, Eli},
  journal={arXiv preprint arXiv:2506.08009},
  year={2025}
}

@article{cui2025self,
  title={Self-forcing++: Towards minute-scale high-quality video generation},
  author={Cui, Justin and Wu, Jie and Li, Ming and Yang, Tao and Li, Xiaojie and Wang, Rui and Bai, Andrew and Ban, Yuanhao and Hsieh, Cho-Jui},
  journal={arXiv preprint arXiv:2510.02283},
  year={2025}
}

@article{yang2025longlive,
  title={Longlive: Real-time interactive long video generation},
  author={Yang, Shuai and Huang, Wei and Chu, Ruihang and Xiao, Yicheng and Zhao, Yuyang and Wang, Xianbang and Li, Muyang and Xie, Enze and Chen, Yingcong and Lu, Yao and others},
  journal={arXiv preprint arXiv:2509.22622},
  year={2025}
}

@inproceedings{ng2022learning,
  title={Learning to listen: Modeling non-deterministic dyadic facial motion},
  author={Ng, Evonne and Joo, Hanbyul and Hu, Liwen and Li, Hao and Darrell, Trevor and Kanazawa, Angjoo and Ginosar, Shiry},
  booktitle={Proceedings of the IEEE/CVF conference on computer vision and pattern recognition},
  pages={20395--20405},
  year={2022}
}

@inproceedings{wang2025diffusion,
  title={Diffusion-based realistic listening head generation via hybrid motion modeling},
  author={Wang, Yinuo and Fan, Yanbo and Wang, Xuan and Yu, Guo and Wang, Fei},
  booktitle={Proceedings of the Computer Vision and Pattern Recognition Conference},
  pages={15885--15895},
  year={2025}
}

@article{zhou2020makelttalk,
  title={Makelttalk: speaker-aware talking-head animation},
  author={Zhou, Yang and Han, Xintong and Shechtman, Eli and Echevarria, Jose and Kalogerakis, Evangelos and Li, Dingzeyu},
  journal={ACM Transactions On Graphics (TOG)},
  volume={39},
  number={6},
  pages={1--15},
  year={2020},
  publisher={ACM New York, NY, USA}
}

@inproceedings{ginosar2019learning,
  title={Learning individual styles of conversational gesture},
  author={Ginosar, Shiry and Bar, Amir and Kohavi, Gefen and Chan, Caroline and Owens, Andrew and Malik, Jitendra},
  booktitle={Proceedings of the IEEE/CVF conference on computer vision and pattern recognition},
  pages={3497--3506},
  year={2019}
}

@article{wan2025wan,
  title={Wan: Open and advanced large-scale video generative models},
  author={Wan, Team and Wang, Ang and Ai, Baole and Wen, Bin and Mao, Chaojie and Xie, Chen-Wei and Chen, Di and Yu, Feiwu and Zhao, Haiming and Yang, Jianxiao and others},
  journal={arXiv preprint arXiv:2503.20314},
  year={2025}
}

@article{zhang2025pretraining,
  title={Pretraining Frame Preservation in Autoregressive Video Memory Compression},
  author={Zhang, Lvmin and Cai, Shengqu and Li, Muyang and Zeng, Chong and Lu, Beijia and Rao, Anyi and Han, Song and Wetzstein, Gordon and Agrawala, Maneesh},
  journal={arXiv preprint arXiv:2512.23851},
  year={2025}
}

@article{tschannen2025siglip,
  title={Siglip 2: Multilingual vision-language encoders with improved semantic understanding, localization, and dense features},
  author={Tschannen, Michael and Gritsenko, Alexey and Wang, Xiao and Naeem, Muhammad Ferjad and Alabdulmohsin, Ibrahim and Parthasarathy, Nikhil and Evans, Talfan and Beyer, Lucas and Xia, Ye and Mustafa, Basil and others},
  journal={arXiv preprint arXiv:2502.14786},
  year={2025}
}

@InProceedings{xie2022vfhq,
      author = {Liangbin Xie and Xintao Wang and Honglun Zhang and Chao Dong and Ying Shan},
      title = {VFHQ: A High-Quality Dataset and Benchmark for Video Face Super-Resolution},
      booktitle={The IEEE Conference on Computer Vision and Pattern Recognition Workshops (CVPRW)},
      year = {2022}
  }

@InProceedings{Chung18b,
  author       = "Chung, J.~S. and Nagrani, A. and Zisserman, A.",
  title        = "VoxCeleb2: Deep Speaker Recognition",
  booktitle    = "INTERSPEECH",
  year         = "2018",
}

@inproceedings{yu2022celebvtext,
  title={{CelebV-Text}: A Large-Scale Facial Text-Video Dataset},
  author={Yu, Jianhui and Zhu, Hao and Jiang, Liming and Loy, Chen Change and Cai, Weidong and Wu, Wayne},
  booktitle={CVPR},
  year={2023}
}

@inproceedings{zhang2021flow,
  title={Flow-Guided One-Shot Talking Face Generation With a High-Resolution Audio-Visual Dataset},
  author={Zhang, Zhimeng and Li, Lincheng and Ding, Yu and Fan, Changjie},
  booktitle={Proceedings of the IEEE/CVF Conference on Computer Vision and Pattern Recognition},
  pages={3661--3670},
  year={2021}
}

@article{li2024openhumanvid,
  title={OpenHumanVid: A Large-Scale High-Quality Dataset for Enhancing Human-Centric Video Generation},
  author={Li, Hui and Xu, Mingwang and Zhan, Yun and Mu, Shan and Li, Jiaye and Cheng, Kaihui and Chen, Yuxuan and Chen, Tan and Ye, Mao and Wang, Jingdong and others},
  journal={arXiv preprint arXiv:2412.00115},
  year={2024}
}

@article{wu2023q,
  title={Q-align: Teaching lmms for visual scoring via discrete text-defined levels},
  author={Wu, Haoning and Zhang, Zicheng and Zhang, Weixia and Chen, Chaofeng and Liao, Liang and Li, Chunyi and Gao, Yixuan and Wang, Annan and Zhang, Erli and Sun, Wenxiu and others},
  journal={arXiv preprint arXiv:2312.17090},
  year={2023}
}

@article{heusel2017gans,
  title={Gans trained by a two time-scale update rule converge to a local nash equilibrium},
  author={Heusel, Martin and Ramsauer, Hubert and Unterthiner, Thomas and Nessler, Bernhard and Hochreiter, Sepp},
  journal={Advances in neural information processing systems},
  volume={30},
  year={2017}
}

@article{unterthiner2018towards,
  title={Towards accurate generative models of video: A new metric \& challenges},
  author={Unterthiner, Thomas and Van Steenkiste, Sjoerd and Kurach, Karol and Marinier, Raphael and Michalski, Marcin and Gelly, Sylvain},
  journal={arXiv preprint arXiv:1812.01717},
  year={2018}
}

@inproceedings{chung2016out,
  title={Out of time: automated lip sync in the wild},
  author={Chung, Joon Son and Zisserman, Andrew},
  booktitle={Asian conference on computer vision},
  pages={251--263},
  year={2016},
  organization={Springer}
}

@misc{simeoni2025dinov3,
  title={{DINOv3}},
  author={Sim{\'e}oni, Oriane and Vo, Huy V. and Seitzer, Maximilian and Baldassarre, Federico and Oquab, Maxime and Jose, Cijo and Khalidov, Vasil and Szafraniec, Marc and Yi, Seungeun and Ramamonjisoa, Micha{\"e}l and Massa, Francisco and Haziza, Daniel and Wehrstedt, Luca and Wang, Jianyuan and Darcet, Timoth{\'e}e and Moutakanni, Th{\'e}o and Sentana, Leonel and Roberts, Claire and Vedaldi, Andrea and Tolan, Jamie and Brandt, John and Couprie, Camille and Mairal, Julien and J{\'e}gou, Herv{\'e} and Labatut, Patrick and Bojanowski, Piotr},
  year={2025},
  eprint={2508.10104},
  archivePrefix={arXiv},
  primaryClass={cs.CV},
  url={https://arxiv.org/abs/2508.10104},
}

@misc{wang2024videoclipxladvancinglongdescription,
      title={VideoCLIP-XL: Advancing Long Description Understanding for Video CLIP Models}, 
      author={Jiapeng Wang and Chengyu Wang and Kunzhe Huang and Jun Huang and Lianwen Jin},
      year={2024},
      eprint={2410.00741},
      archivePrefix={arXiv},
      primaryClass={cs.CL},
      url={https://arxiv.org/abs/2410.00741}, 
}

@misc{tu2025stableavatarinfinitelengthaudiodrivenavatar,
      title={StableAvatar: Infinite-Length Audio-Driven Avatar Video Generation}, 
      author={Shuyuan Tu and Yueming Pan and Yinming Huang and Xintong Han and Zhen Xing and Qi Dai and Chong Luo and Zuxuan Wu and Yu-Gang Jiang},
      year={2025},
      eprint={2508.08248},
      archivePrefix={arXiv},
      primaryClass={cs.CV},
      url={https://arxiv.org/abs/2508.08248}, 
}

@article{ding2025mtvcrafter,
  title={Mtvcrafter: 4d motion tokenization for open-world human image animation},
  author={Ding, Yanbo and Hu, Xirui and Guo, Zhizhi and Zhang, Chi and Wang, Yali},
  journal={arXiv preprint arXiv:2505.10238},
  year={2025}
}

@article{raffel2020exploring,
  title={Exploring the limits of transfer learning with a unified text-to-text transformer},
  author={Raffel, Colin and Shazeer, Noam and Roberts, Adam and Lee, Katherine and Narang, Sharan and Matena, Michael and Zhou, Yanqi and Li, Wei and Liu, Peter J},
  journal={Journal of machine learning research},
  volume={21},
  number={140},
  pages={1--67},
  year={2020}
}

@article{xu2025qwen3,
  title={Qwen3-omni technical report},
  author={Xu, Jin and Guo, Zhifang and Hu, Hangrui and Chu, Yunfei and Wang, Xiong and He, Jinzheng and Wang, Yuxuan and Shi, Xian and He, Ting and Zhu, Xinfa and others},
  journal={arXiv preprint arXiv:2509.17765},
  year={2025}
}

@article{zhao2023pytorch,
  title={Pytorch fsdp: experiences on scaling fully sharded data parallel},
  author={Zhao, Yanli and Gu, Andrew and Varma, Rohan and Luo, Liang and Huang, Chien-Chin and Xu, Min and Wright, Less and Shojanazeri, Hamid and Ott, Myle and Shleifer, Sam and others},
  journal={arXiv preprint arXiv:2304.11277},
  year={2023}
}

@misc{schuhmann2022laion5bopenlargescaledataset,
      title={LAION-5B: An open large-scale dataset for training next generation image-text models}, 
      author={Christoph Schuhmann and Romain Beaumont and Richard Vencu and Cade Gordon and Ross Wightman and Mehdi Cherti and Theo Coombes and Aarush Katta and Clayton Mullis and Mitchell Wortsman and Patrick Schramowski and Srivatsa Kundurthy and Katherine Crowson and Ludwig Schmidt and Robert Kaczmarczyk and Jenia Jitsev},
      year={2022},
      eprint={2210.08402},
      archivePrefix={arXiv},
      primaryClass={cs.CV},
      url={https://arxiv.org/abs/2210.08402}, 
}

@misc{yang2025qwen3technicalreport,
      title={Qwen3 Technical Report}, 
      author={An Yang and Anfeng Li and Baosong Yang and Beichen Zhang and Binyuan Hui and Bo Zheng and Bowen Yu and Chang Gao and Chengen Huang and Chenxu Lv and Chujie Zheng and Dayiheng Liu and Fan Zhou and Fei Huang and Feng Hu and Hao Ge and Haoran Wei and Huan Lin and Jialong Tang and Jian Yang and Jianhong Tu and Jianwei Zhang and Jianxin Yang and Jiaxi Yang and Jing Zhou and Jingren Zhou and Junyang Lin and Kai Dang and Keqin Bao and Kexin Yang and Le Yu and Lianghao Deng and Mei Li and Mingfeng Xue and Mingze Li and Pei Zhang and Peng Wang and Qin Zhu and Rui Men and Ruize Gao and Shixuan Liu and Shuang Luo and Tianhao Li and Tianyi Tang and Wenbiao Yin and Xingzhang Ren and Xinyu Wang and Xinyu Zhang and Xuancheng Ren and Yang Fan and Yang Su and Yichang Zhang and Yinger Zhang and Yu Wan and Yuqiong Liu and Zekun Wang and Zeyu Cui and Zhenru Zhang and Zhipeng Zhou and Zihan Qiu},
      year={2025},
      eprint={2505.09388},
      archivePrefix={arXiv},
      primaryClass={cs.CL},
      url={https://arxiv.org/abs/2505.09388}, 
}

@misc{cheng2026unisonharmonizingmotionspeech,
      title={Unison: Harmonizing Motion, Speech, and Sound for Human-Centric Audio-Video Generation}, 
      author={Shihao Cheng and Jiaxu Zhang and Quanyue Song and Shansong Liu and Zhizhi Guo and Xiaolei Zhang and Chi Zhang and Xuelong Li and Zhigang Tu},
      year={2026},
      eprint={2605.08729},
      archivePrefix={arXiv},
      primaryClass={cs.CV},
      url={https://arxiv.org/abs/2605.08729}, 
}
\end{document}


\title{Appendix} 

\titlerunning{InteractiveAvatar}


\author{Quanyue Song\inst{1,2}$^{\ast\dagger}$ \and
Yishan He\inst{2}$^{\dagger}$ \and
Yanfei Zhang\inst{2} \and
Shihao Cheng\inst{2,3}$^{\ast}$ \and
Zhixiang He\inst{2} \and
Zhizhi Guo\inst{2}\textsuperscript{\Letter} \and
Chi Zhang\inst{4} \and
Xuelong Li\inst{4} \and
Caigui Jiang\inst{1}\textsuperscript{\Letter}
}

\authorrunning{Q.~Song et al.}

\institute{State Key Laboratory of Human-Machine Hybrid Augmented Intelligence, Institute of Artificial Intelligence and Robotics, Xi'an Jiaotong University, China \and
China Telecom Artificial Intelligence Technology (Beijing) Co., Ltd., China \and
State Key Laboratory of Information Engineering in Surveying, Mapping and Remote Sensing, Wuhan University, China \and
Institute of Artificial Intelligence (TeleAI), China Telecom, China
}

\maketitle

\begingroup
\renewcommand\thefootnote{}\footnotetext{$^{\ast}$ Work done during an internship at China Telecom Artificial Intelligence Technology (Beijing) Co., Ltd.}
\renewcommand\thefootnote{}\footnotetext{$^{\dagger}$ Equal contribution.}
\renewcommand\thefootnote{}\footnotetext{\textsuperscript{\Letter} Corresponding author.}
\endgroup

\section{Training and Inference}
Our training pipeline consists of four stages, progressively equipping the model with audio-driven control and ultimately distilling it into a real-time generation model. Stage 1 initializes audio-driven generation, Stage 2 learns LSVM compression, Stage 3 (ODE initialization with cache switching) and Stage 4 are necessary DMD-specific training stages, where Self-Forcing DMD is applied to achieve stable few-step rollout. This progression separates capability acquisition, memory learning, ODE initialization, and real-time distillation.

The stage one is audio-driven capability initialization. We first build the audio-driven generation ability based on the Wan 2.2 5B~\cite{wan2025wan} backbone. Following~\cite{kong2025let}, the input audio is encoded by a Wav2Vec~\cite{prajwal2020lip} encoder to obtain speech features. These audio features are injected into the DiT backbone through cross-attention layers, enabling the model to condition generation on speech signals. During this stage, only the cross-attention parameters are trained while the backbone weights remain frozen.

In the second stage, we train the memory compression module. For each training sample, we select a video sequence and divide its historical context into short-term and long-term memories. Frames within 5 seconds of a selected reference timestamp are treated as short-term memory, while frames earlier than 5 seconds are treated as long-term memory. Since compression is performed in the video latent space, the short-term memory compresses all video latents within the 5-second window. For the long-term memory, we randomly sample latent tokens from frames earlier than 5 seconds to match the predefined long-term memory capacity. A target latent is then selected either from the short-term memory or from a neighboring frame of one sampled long-term latent. This latent is used as the prediction target. In this stage, both the memory compression module and the DiT backbone are jointly fine-tuned.

\begin{figure*}[t]
\centering
\includegraphics[width=\textwidth, trim=20 270 20 5, clip]{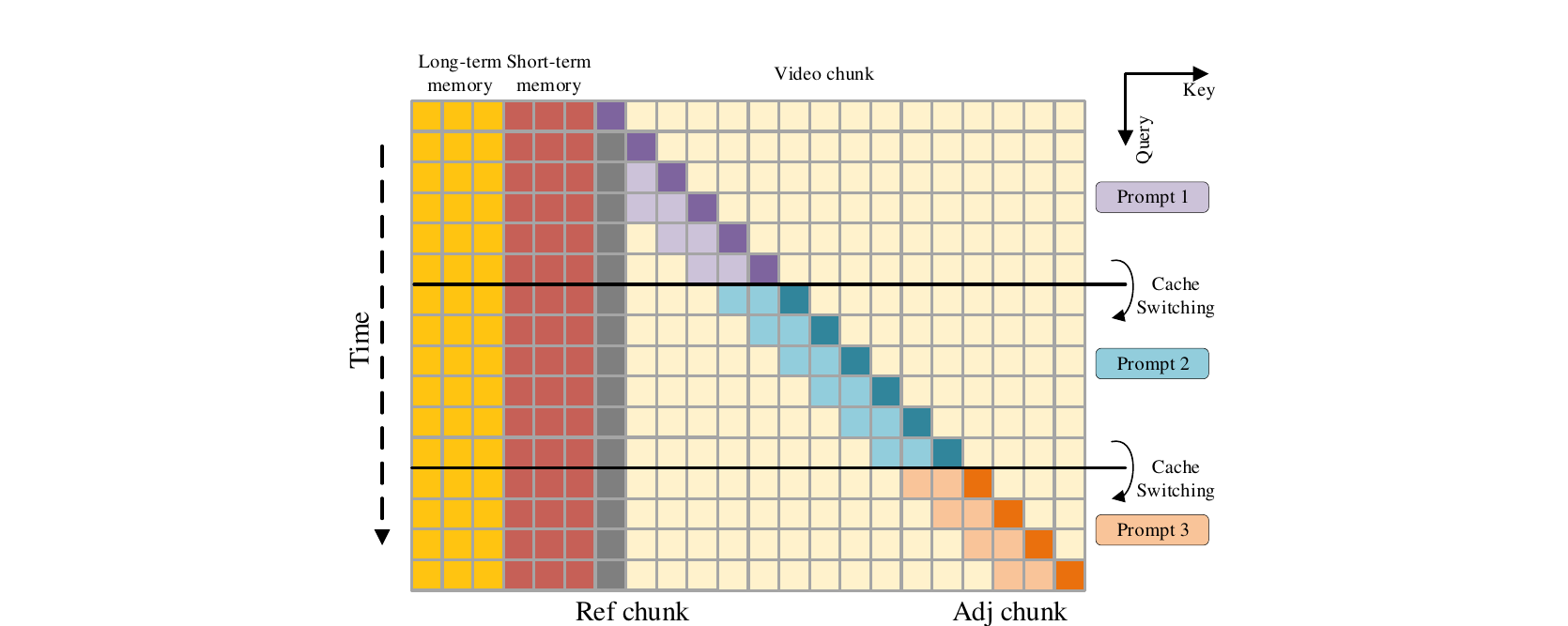} 
\caption{Causal attention mask for our InteractivateAvatar. Each generated chunk attends to the reference chunk, the two preceding adjacent chunks, and both short-term and long-term memory latents to ensure temporal consistency.}
\label{fig_memory_mask}
\end{figure*}

The stage three is student Initialization. We modify the attention mask of the model to a specialized causal attention structure and treat it as a student model (Fig.~\ref{fig_memory_mask}). Training samples are generated using a teacher model. We then perform ODE training to fine-tune the initialized student model. The designed attention mask ensures that each generated chunk attends to its two preceding neighboring chunks, the first chunk generated from the initial frame, and the compressed memory representations.

The stage four is autoregressive distillation training~\cite{huang2025self}. Finally, we apply Self-Forcing DMD training to perform autoregressive distillation, enabling the model to stably generate long sequences under streaming inference settings.

During inference, we first initialize the short-term and long-term memories. The model generates the first chunk conditioned on the reference frame of the character. After the first chunk is generated, the second chunk is produced by attending to the first chunk. For subsequent generation, each new chunk attends to the two preceding neighboring chunks, the first chunk, and the compressed short-term and long-term memories. At every inference step, both memory modules are updated, and the necessary KV caches are stored to avoid redundant computation.

\section{Parameter Settings}

During both training and inference, each chunk consists of three video latents, and the generation of a chunk attends to the two adjacent chunks as context. Increasing the number of video latents within each chunk or increasing the number of referenced chunks provides richer contextual information but significantly increases computational cost, making real-time inference difficult to achieve. Conversely, reducing the number of latents per chunk or the number of referenced chunks limits the available context, which can degrade temporal consistency in the generated video. In our design, we find a practical balance between generation quality and inference efficiency.

To further enable real-time generation, we distill the model to 3 diffusion steps. We also experimented with distillation to 2 steps, However, this resulted in a noticeable degradation in visual quality, particularly causing significant changes in the identity of the characters. For the output resolution, we choose the highest resolution that still supports real-time inference, generating videos at 576p (1024 × 576 with a 16:9 aspect ratio or 896 × 672 with a 4:3 aspect ratio or 768 x 768 with a 1:1 aspect ratio).

\begin{table}[tb]
\caption{Hyperparameter sensitivity.}
\label{tab:Hyperparameter_sensitivity}
\centering
\begin{tabular}{lccccccc}
\toprule
Setting & IQA$\uparrow$ & FVD$\downarrow$ & OBJ$\uparrow$ & ID$\uparrow$& TV$\uparrow$  & FPS$\uparrow$  &TTFF(s)$\downarrow$\\
\midrule
One adjacent chunks & 3.54 & 847.7 & 82.1 & 4.47 & \textbf{25.95} & \textbf{29.64} & \textbf{2.5}\\
0.5x memory(15 latents)& 3.85 & 731.9 & 81.7 & 4.49 & 25.89 & \underline{28.17} & \underline{2.5}\\
2x memory(60 latents)& \textbf{3.89} & \underline{711.6} & \textbf{85.5} & \underline{4.51} & 25.93 & 21.52 & 2.8\\
Our settings & \underline{3.87} & \textbf{701.4} & \underline{85.2} & \textbf{4.51} & \underline{25.93} & 26.68 & 2.6\\
\bottomrule
\end{tabular}
\end{table}

For the memory design, we compress a 5-second (120 frames, 30 latents) video segment as short-term memory, while the long-term memory is assigned the same capacity as the short-term memory. In Tab.~\ref{tab:Hyperparameter_sensitivity}, we compare the performance of different parameter configurations. Our setup achieves competitive results while maintaining real-time performance and acceptable Time-to-First-Frame (TTFF). Additionally, we adopt two adjacent chunks, as we observe that this configuration produces visually smoother video outputs. The spatial dimensions of the latents are compressed by a factor of 4× in both height and width, while the temporal dimension is preserved without compression. Under this configuration, each memory latent corresponds to 1024/16/4 × 576/16/4 = 144 tokens at 576p resolution. R is dynamically updated according to Eq.10 in the main text, using the minimum R from the previous Key-Frame Selection step. This design maintains efficient generation speed while preserving as much temporal and visual consistency as possible.

Our Reasoning-Reaction Module consists of two main modules. The audio-to-audio model utilizes the Doubao framework for A2A generation, while the reasoning module employs Qwen3 4B~\cite{yang2025qwen3technicalreport} or Qwen3 Omni~\cite{xu2025qwen3} for action and context reasoning. To accelerate inference, we adopt pipeline parallelism, distributing the DiT backbone and VAE across two H100 GPUs. Communication between the GPUs is implemented via NCCL-based NVLink, enabling efficient feature transfer. Finally, the generated video stream is delivered to the client in real time using a WebSocket-based streaming pipeline.

\section{Cache Switching}

\begin{figure*}[t]
\centering
\includegraphics[width=\textwidth, trim=20 170 10 5, clip]{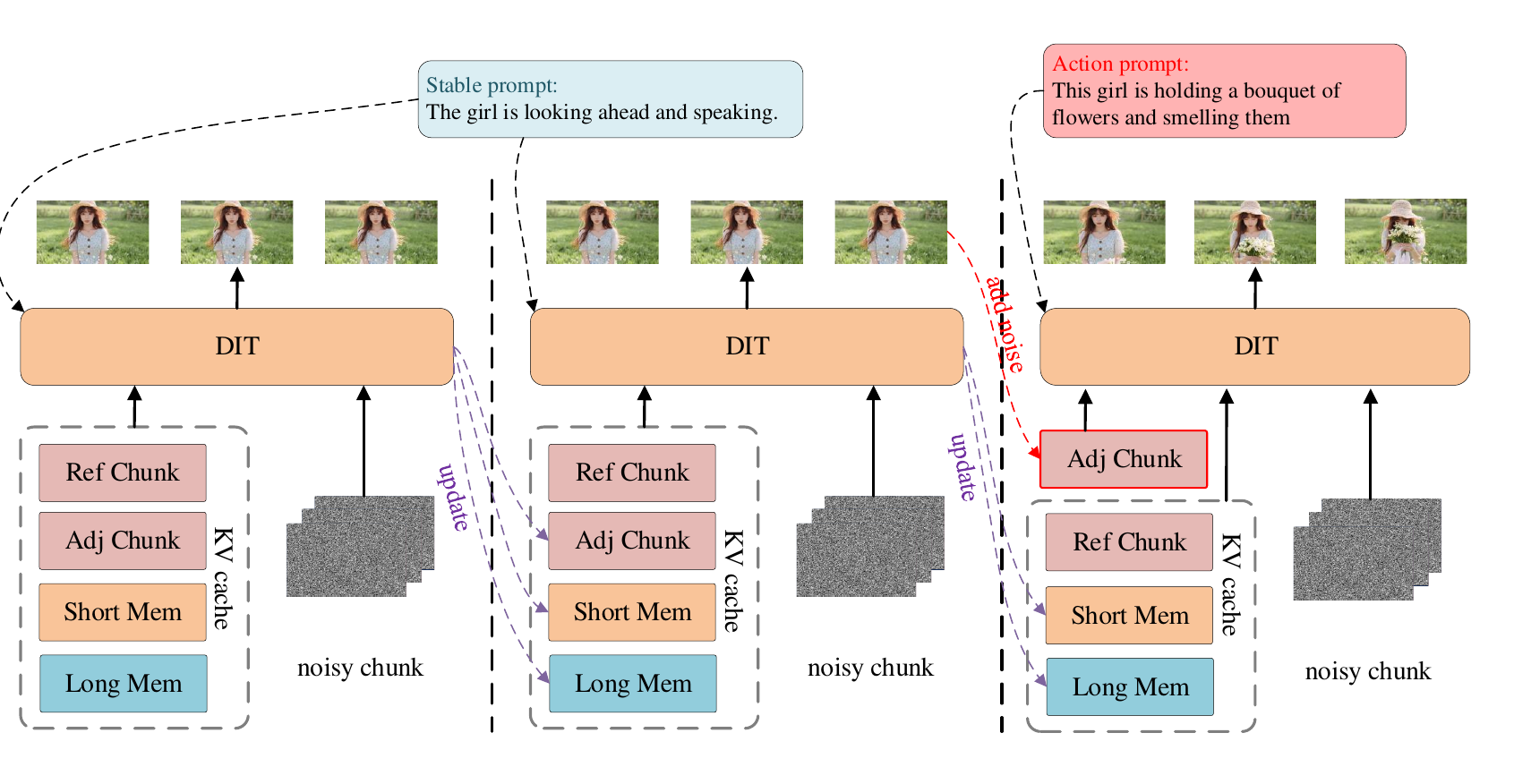} 
\caption{Cache Switching Mechanism. Upon receiving a new action prompt, previously generated frames are re-noised and their latent KV are recomputed, enabling the model to rapidly adapt to the updated prompt.}
\label{fig_cache_switching}
\end{figure*}

To reduce the response latency of generated videos to changes in the input prompt, we introduce a cache switching mechanism. As illustrated in Fig.~\ref{fig_cache_switching}, when the prompt used for generating the current chunk is the same as that of the previous step, we reuse the cached KV representations of the reference chunk, neighboring chunks, and the short-term and long-term memories to avoid redundant computation.

When the prompt used for the current chunk differs from the previous one, we discard the cached KV representations of the neighboring chunks to prevent information associated with the previous prompt from interfering with the current generation. However, the KV caches corresponding to the reference chunk and the memory modules remain unchanged.

To recompute the KV representations, we add noise to the already generated neighboring video latents to match the corresponding noise level. These re-noised latents are concatenated with the current noisy latent to be denoised and other conditioning inputs, after which the KV representations are recomputed and cached. This process enables rapid adaptation to prompt changes while maintaining temporal continuity in the generated video.

\section{Training Data}
Our training data consists primarily of publicly available datasets along with a portion of proprietary data. Detailed statistics are provided in the main paper. For all data sources, we apply a unified data processing pipeline.

First, we evaluate the aesthetic score~\cite{schuhmann2022laion5bopenlargescaledataset} of each video and perform shot boundary detection to ensure visual clarity and temporal continuity. We further apply face detection and lip-sync~\cite{chung2016out} verification to filter out videos without human subjects or with mismatched audio–visual synchronization.

During the captioning stage, we use Gemini to identify action-related events of the character. For each detected event, an action prompt is assigned at the beginning of the event, while a stable prompt describing the current state of the character is assigned at the end of the event. Through this process, each video is associated with a sequence of prompts, each annotated with a timestamp and labeled as either an action prompt or a stable prompt.

\section{Identity metric}
DINOv3 alone may be insufficient for strict ID evaluation. We use it because avatar videos involve full-body appearance and clothing beyond facial identity. We uniformly sampled frames from the generated videos and evaluated face similarity using ArcFace. As shown in the  Tab.~\ref{tab:Identity_metric}, our method achieves competitive scores on this metric.

\begin{table}[tb]
\caption{Identity metric.}
\label{tab:Identity_metric}
\tiny
\centering
\begin{tabular}{lcccccccc}
\toprule
Metrics & StableAvatar & OmniAvatar & HYAvatar & Hallo3 & EchoMimicV3  & WanS2V  & LiveAvatar & Ours\\
\midrule
ID (ArcFace) $\uparrow$
& 0.72 & 0.75 & 0.81 & 0.70 & 0.81 & 0.78 & \textbf{0.84} & \underline{0.82} \\
\bottomrule
\end{tabular}
\end{table}

\section{More examples}
Below we present more examples demonstrating user intent understanding and interactions between users and the avatar (Fig. ~\ref{fig_example_1} and Fig.~\ref{fig_example_2}). Please refer to the videos in the supplementary material.

\begin{figure*}[t]
\centering
\includegraphics[width=\textwidth, trim=20 350 100 5, clip]{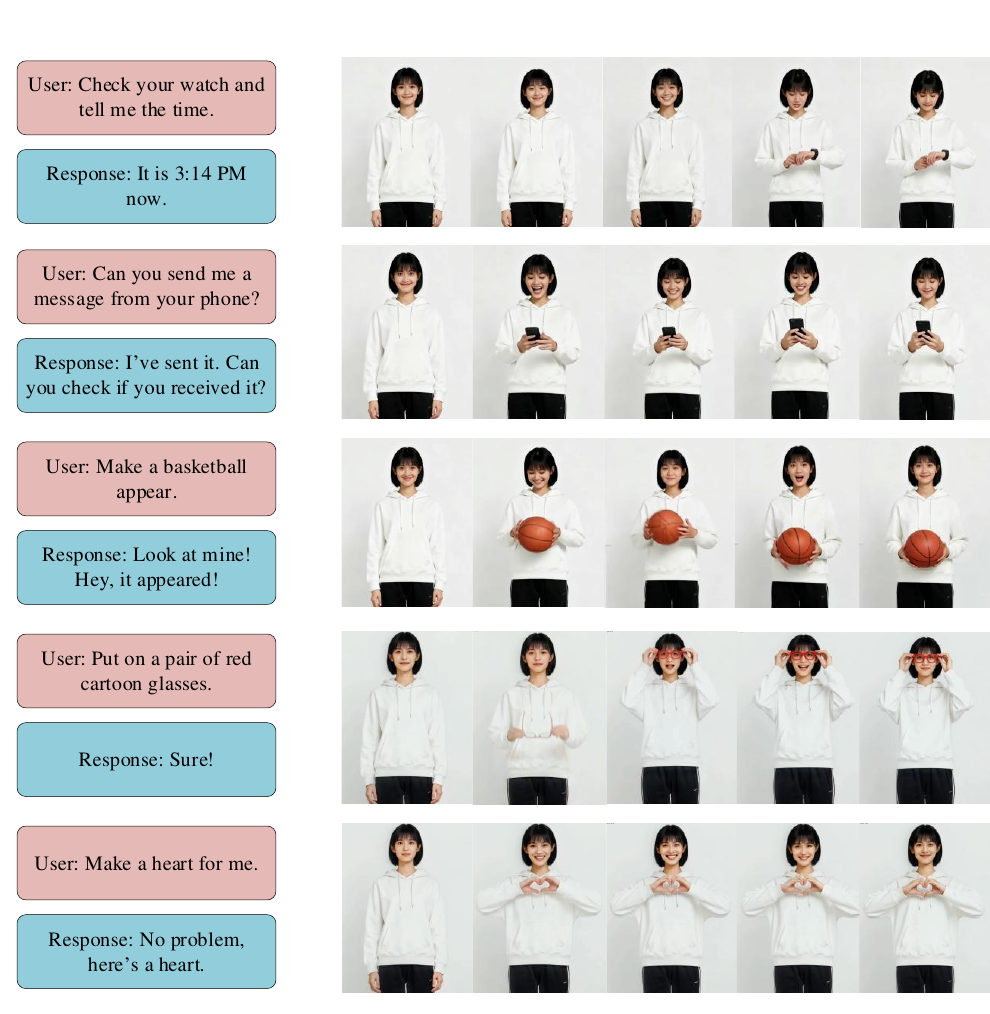} 
\caption{More examples: user intent understanding.}
\label{fig_example_1}
\end{figure*}

\begin{figure*}[t]
\centering
\includegraphics[width=\textwidth, trim=5 5 5 5, clip]{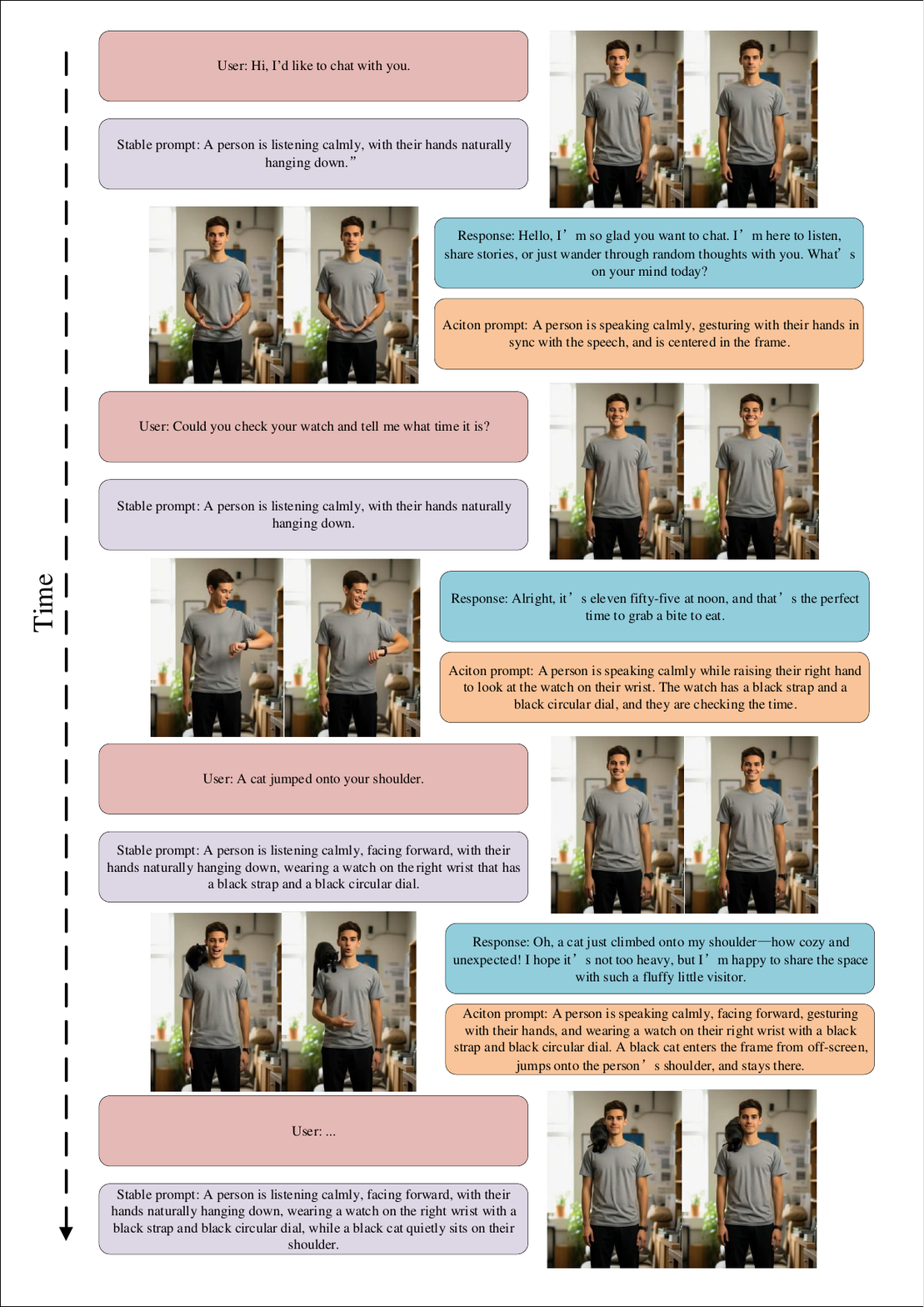} 
\caption{More examples: the interactions between users and the avatar.}
\label{fig_example_2}
\end{figure*}

\clearpage

\bibliographystyle{splncs04}
\bibliography{main}